\documentclass[10pt,journal,compsoc]{IEEEtran}
\ifCLASSOPTIONcompsoc
  \usepackage[nocompress]{cite}
\else
  \usepackage{cite}
\fi
\ifCLASSINFOpdf
\else
\fi
\usepackage{times}
\usepackage{epsfig}
\usepackage{graphicx}
\usepackage{amsmath}
\usepackage{amssymb}
\usepackage{times}
\usepackage{epsfig}
\usepackage{graphicx}
\usepackage{amsmath}
\usepackage{amssymb}
\usepackage{algpseudocode,algorithm,algorithmicx}
\usepackage{subfigure}
\usepackage{comment}
\usepackage{epsfig}

\usepackage{epstopdf}
\usepackage{booktabs, multicol, multirow}
\usepackage{setspace}

\begin{document}
\title{Attribute Guided Sparse Tensor-Based Model for Person Re-Identification}
% lists the Biometrics Council journals use for "first footnote" author

\author{Fariborz Taherkhani, Ali Dabouei, Sobhan Soleymani, Jeremy Dawson, and Nasser M. Nasrabadi\\
Lane Department of Computer Science and Electrical Engineering \\ West Virginia University}


\markboth{Journal of \LaTeX\ Class Files,~Vol.~14, No.~8, August~2015}%
{Shell \MakeLowercase{\textit{et al.}}: Bare Demo of IEEEtran.cls for IEEE Journals}

% Biometrics Council journal papers as those publications place the marks
% or like this to get the Biometrics Council new two part style.
%\hspace{\columnsep}\makebox[\columnwidth]{Published by the IEEE Biometrics Council\hfill}}
% column for its text to clear the IEEEpubid mark (Biometrics Council jorunal


% for Biometrics Council papers, we must declare the abstract and index terms
\IEEEtitleabstractindextext{%
\begin{abstract}
Visual perception of a person is easily influenced by many factors such as camera parameters, pose and viewpoint variations. These variations make person Re-Identification (ReID) a challenging problem. Nevertheless, human attributes usually stand as  robust visual properties to such variations. In this paper, we propose a new  method to leverage  features from human attributes for person ReID. Our model uses a tensor to non-linearly  fuse  identity  and  attribute features, and then forces the parameters of the tensor in the  loss function to generate  discriminative fused features for ReID. Since tensor-based methods usually contain a large number of parameters, training all of these parameters becomes very slow, and the chance of overfitting increases as well. To address this issue, we propose two new techniques based on Structural Sparsity Learning (SSL) and Tensor Decomposition (TD) methods to create  an  accurate  and  stable  learning  problem. We conducted experiments on several standard pedestrian datasets, and experimental results indicate that our tensor-based approach significantly improves person ReID baselines and also outperforms state of the art methods.
\end{abstract}

}

\maketitle

\IEEEdisplaynontitleabstractindextext

% \fi
\IEEEpeerreviewmaketitle

%\doublespacing

\IEEEraisesectionheading{\section{Introduction}\label{sec:introduction}}

The task of person ReID involves matching people across multiple cameras, or across time within a single camera \cite{xu2014person}, and the task of attribute prediction is to decide the presence of a set of attributes from a given image \cite{kumar2009attribute}. Research in person ReID has attracted much attention in computer vision and biometrics communities due to its usefulness in variety of  applications such as robotics, human-computer interaction, video surveillance,  etc.
 Despite years of effort, person ReID still remains a challenging problem \cite{lin2019improving,alemu2019deep,zheng2019pyramidal,chen2019abd,chen2020salience,luo2020generalizing,porrello2020robust,gao2020pose} due to 1) large variations in human pose across time and space, 2) background clutter
and occlusions, and 3) variations in camera viewpoints and lighting conditions. Even though the appearance of a person significantly changes under these variations, attributes that possess high-level semantic context with respect to the person remain comparatively consistent and stable \cite{layne2012person,zhao2019attribute,guo2019beyond}. 

%Therefore, we can leverage these robust visual features as a source of complementary information to improve the challenge of person ReID performance.

Most of the person ReID methods that are based on Convolutional Neural Networks (CNN) use the global descriptors because the typical CNN features used for these methods are usually the descriptors which represent the global structure of a person \cite{hou2020iaunet}. However, the features used in attribute-based methods usually represent the local structures of a person \cite{han2018attribute,wang2018transferable,tay2019aanet,zhao2019attribute,watson2020person}. We speculate that local attribute features provide complementary information along with the typical global CNN features. Thus, these complementary aspects can be  leveraged to improve the performance of the person ReID. In other words, a  person ReID method might not discriminate the slight differences between two identities  in cases where their appearances are similar to each other. However, a person ReID algorithm can make a more accurate decision by considering the details provided by their attributes. For example, a person ReID method might not be able to discriminate identities wearing similar red and white clothes. However, attributes such as 'man wearing hat and bag' might provide  some complementary information  to help the person ReID algorithm to distinguish them accurately. For example,
Su \textit{et al.} \cite{su2015multi} propose a multi-task learning framework with a low-rank attribute embedding for person ReID. The low-rank attribute embedding learns to transform the original binary attributes to a continuous attribute space where incorrect and incomplete attributes are refined to represent identities in a more efficient way.

Attributes have  been used in many person ReID methods \cite{layne2012person,layne2014re,wang2018mancs,lin2019improving,zhao2019attribute,han2019attribute,guo2019beyond,watson2020person, zhang2020person,schumann2017person} and biometric applications such as face recognition \cite{wang2017multi,hu2017attribute,taherkhani2018deep}. In most of these methods,  attributes have provided auxiliary information for person ReID. For example, methods in \cite{layne2014attributes,layne2014re} and \cite{layne2012person} combine low-level image descriptors with attribute information to create a complement  set of features for person ReID. Khamis \textit{et al}. \cite{khamis2014joint} 
learn a discriminative transform  to project the input images to a joint appearance-attribute subspace where interaction between the attributes and the appearance of the person is leveraged for matching.
Matsukawa \textit{et al}. \cite{matsukawa2016person}  enhance the CNN features by performing a fine-tuning step on a pedestrian-attribute dataset. Here, different attribute prediction losses are added to pedestrian classification loss  to differentiate more subtle information and  obtain more discriminative features. Generally, most of the attribute-based person ReID methods usually use image pairs or triplets to train their CNN framework. 

%and several dataset have been designed for this purpose. For example, Deng \textit{et al}. \cite{deng2014pedestrian} and  Li \textit{et al}. \cite{li2016richly}  released two large scale pedestrian-attribute namely PETA and RAP datasets.

%Attribute has been exploited in biometric application specially for the face recognition challenges \cite{taherkhani2018deep, wang2017multi}. For example, Wang et al. \cite{wang2017multi} propose an attribute-constrained face recognition model for joint facial attributes prediction and face recognition. In this model, the parameters of the network are first updated for attributes prediction and then same network is fine-tuned for face recognition.  Ranjan et al. \cite{ranjan2017all} add other face related tasks to improve overall performance. Their model is a single multi-task CNN network for simultaneous face detection, face alignment, pose estimation, gender recognition, smile detection, age estimation and face recognition. 

Tensor Modeling (TM) has a long history in addressing computer vision and machine learning problems. TM can be used to  recover latent factors in complex data and arrange a natural model to handle  the intrinsically complicated structure of the visual data and their multi-modal aspects. TM was first utilized  for face recognition \cite{vasilescu2002multilinear}, human motion recognition \cite{vasilescu2002human} and person ReID \cite{zhang2019tensor}. Recently, TM  has  successfully contributed  to training deep neural networks \cite{kiros2014multimodal,kiros2014unifying,yang2016deep},  revealing high-order relations in the data \cite{duchenne2011tensor}, unsupervised learning of latent variable models \cite{anandkumar2014tensor}, and justifying some of their theoretical aspects. There  are  basically two types of TM techniques, linear and multi-linear tensor data modeling. In contrast to  linear tensor methods (e.g., CANDECOMP/Parafac),   Multi-linear tensor methods (e.g., Tucker/M-mode SVD) are more appropriate for multi-modal data analysis \cite{hou2019deep} where the data, such as face images, can be directly represented  by the color values in the pixel domain, or information related to soft-biometrics such as gender, nose shape,  etc.

Fusion approaches for classification and verification tasks are roughly divided into feature-level and score-level categories. In score-level fusion,  similarity scores obtained from each modality are fused by using a simple voting, or by stacking another multi-class linear classifier \cite{simonyan2014two}. In feature-level fusion, features  are fused either by subspace learning or simple feature aggregation. For subspace learning methods, the features from different modalities are first concatenated together and then they are projected to a subspace such that feature of each modality provides complementary information for the other one. The projection can be either in a supervised fashion such as  Linear Discriminant Analysis (LDA) \cite{belhumeur1997eigenfaces} or Locality Preserving Projections (LPP) \cite{he2005face},  or in an unsupervised fashion such as  Bilinear Models (BLM) \cite{tenenbaum1997separating}, or Canonical Correlational Analysis (CCA) \cite{shawe2004kernel}. For aggregation methods, features are usually fused by element wise averaging, product, or concatenation \cite{park2016combining}. 

  In this paper, we propose a  tensor-based model which leverages the person attributes for identification. In this work, the tensor is an operator which conducts two tasks jointly. The first task is to fuse information from the two sources of  information (i.e., features of the attributes and identity) while the second task is to learn discriminative features from the identities based on the classification and contrastive losses. In the first task, the tensor non-linearly fuses the features of the identity and attributes. In the second task, the tensor aims to bring the features of the genuine pairs close to each other while pushing away the features of the imposter pairs to increase the discrimination of the fused features. Since tensor-based models usually contain a large number of parameters, training all parameters of the tensor becomes very slow and the chance of  overfitting increases as well. To address this issue, we propose two new techniques in our method to reduce the total number of  parameters during the training step. The first technique uses a Structural Sparsity Learning (SSL) method which is applied to the total loss function to regularize the parameters of the tensor, while the second technique  directly uses a Tensor Decomposition (TD) method within the model to regularize the parameters of the tensor during the training.
  
\section{Related Work}
Recently, inspired by  CNNs which have provided promising results for various problems in computer vision, ReID based on CNN methods
have attracted significant
attention \cite{zheng2019re,chen2019mixed, dai2019batch,tay2019aanet,hou2019interaction, zhao2020not,chen2020salience}. In general, previous works on person ReID mostly focus on either designing feature representations which are not sensitive to the view-point  \cite{liu2018pose,sun2019perceive,sarfraz2018pose,porrello2020robust},  learning an efficient distance-metric \cite{kalayeh2018human,zheng2018pedestrian,ali2018maximum,wang2019learning,luo2019bag,ahmed2020camera}, or  methods which consider both factors \cite{sun2017svdnet,wang2018mancs,sun2018beyond,si2018dual,jin2020style}.

CNN-based person ReID methods are roughly divided into two categories: 1) deep representation learning, and 2) deep metric learning. The first category \cite{ geng2016deep, zheng2017discriminatively, zheng2017person} has become progressively well-known in the person ReID research community due to their efficacy. An example of this category is presented in Xiao \textit{et al}. \cite{xiao2016learning} which designs
a pipeline to learn deep feature representations from multiple
domains using CNN. The method proposes a domain guided dropout algorithm to enhance the feature learning process. Methods presented in \cite{xiaoend}, and \cite{zheng2017person} also provide  an end-to-end deep learning framework to train
pedestrian detection and person
ReID jointly with the goal of improving the overall person ReID performance. Inspired by the method presented in \cite{sun2014deep}, methods \cite{zheng2017discriminatively,geng2016deep}  combine identification loss and verification loss to learn more discriminative descriptors for person ReID. In a second category of approaches in \cite {li2014deepreid, ahmed2015improved, chen2017multi},  deep metric learning, image pairs or triplets are usually given to the network. These methods usually include spatial constraints when a similarity learning process is conducted.  For example, Varior \textit{et al}. \cite{li2014deepreid} employ a gating function after each convolutional layer to consider 
the differences in fine common local patterns between parts across
the whole CNN network. Chen \textit{et al}. \cite{chen2017multi} propose a multi-task loss function in which both ranking loss and verification loss are taken into account.  Both of these losses are optimized  simultaneously for person
ReID. Deep metric learning methods are typically trained  properly on small datasets. However, they might not be a perfect method for training on  large-scale person ReID datasets.
\begin{comment}
\section{Fusing Attributes and Identity Features}

In this section, we define a tensor-based model which predicts people identities.  This model fuses extracted features from human attributes (HA) and identity (ID) and then employs the output for ReID. Each set of features (i.e, HA and ID features)  are extracted from two separate CNN networks; one network is used to extract HA features while the other one is used to extract ID features. Therefore, our model takes output of these networks as an input to fuse them. Depending on the problem and available auxiliary information, this model can be extended to fuse more than two sets of features; however, in this paper, we only concentrate on HA and ID features for the ReID problem. We start with a linear model which performs on single set feature (i.e, ID feature) for ReID problem and then use tensor operations to extend it to multiple sets of features (i.e, ID and HA features). This model fuses features non-linearly and leverages it for person identity prediction. In the next section we formulate ReID problem using our model for two different cases (single and multiple features) and also provide some background information about tensor operations.
\end{comment}
 \begin{figure}[t]
\centering
\includegraphics [scale=.4]{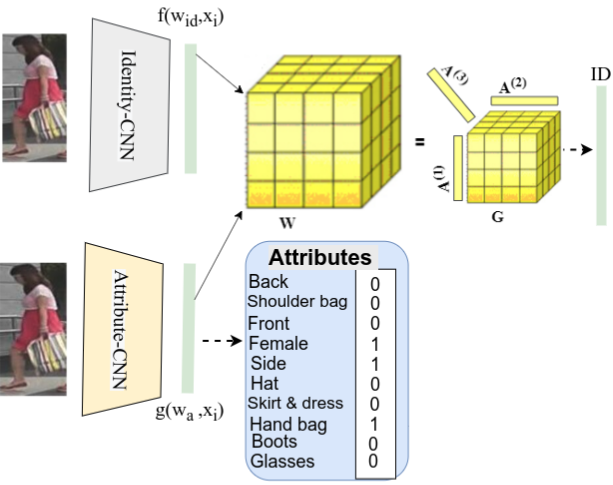}
\caption{Attribute based Tensor fusion diagram for Person ReID.}\label{fig1} 
\end{figure}
\section{Problem Formulation and Background}
Here, we formulate the ReID problem for the case where a set of identities and attributes features are available during the training.
\begin{comment}
\textbf{Identity Feature:} Assume that $X=\{x_1,x_2,...,x_N\}$ are the training images which are given to the CNN network \textit{f} with the weighting parameter $W_{id}$ (Fig. \ref{fig:results}). The \textit{N} is number of the training samples and $f(x_i,W_{i})$ is a \textit{D}-dimensional feature vector which is extracted from a CNN network for the given image $x_i$. This feature vector  is obtained from the feature maps which are captured from the last convolutional layers by flattening and concatenating all of them. Here, we consider the ReID with the multi-class classification problem setting. The label space has \textit{C} unique number of
classes (number of identities), so each sample in $Y=\{y_1,y_2,...,y_N\}$ is corresponded by a \textit{C}-dimensional one-hot encoding label vector. By considering a linear model for  ReID problem,  labels $ \tilde{Y}=\{\tilde{y_1}, \tilde{y_2},...,  \tilde{y_N}\}$ are predicted by a simple dot product between feature vector obtained from the $f$ network and  parameters of the linear classifier $W_c$ as follows: $$ \tilde{y_j}=W_c {f^\top (x_j,W_{id})}.$$ where $W_c$ and $f^\top (x_j,W_{id}) $ are a row vector.
\end{comment}

%A simple approach for fusing is to  concatenate  $[f(x_j,W_{id}), g(x_j,W_a)]$ together and directly plug it back to the Eq (). This chose, however, is a linear fusion method and disregards the interaction between attribute and identity features. 
\textbf{Identity and Attributes Features:}  consider  that, other than the identity features,  attribute features are also available, and we aim to leverage them for the ReID problem (Fig. \ref{fig1}). Assume that  $A$-dimensional vectors, $ g(x_i,w_a)$  and $f(x_i,w_{i})$, are the  attribute and identity feature vectors extracted by two CNN networks with  parameters $w_a$ and $w_i$  for the given image $x_i$, respectively. Thus, for the given image $x_j$, we have a  pair feature vectors $ \{f (x_j,w_{i}), g(x_j,w_a)\} $ which represents identity and attribute information, respectively. A simple approach for leveraging the attribute features is to fuse them by concatenating  $[f(x_j,w_{i}), g(x_j,w_a)]$ together and directly use it for the ReID. This solution, however, is a linear fusion method and disregards the interaction between attribute and identity features. Here, we develop a non-linear fusion method with the following formulation to consider the interaction between the attribute and identity features in the model. We present the ReID problem by using the attribute and identity features as follows: 
\begin{equation}
    \tilde{y_j}=\textbf{W}\times_1 f(x_j,w_{i})\times_3 g(x_j,w_a),
\end{equation}
    
where $\textbf{W}$ is a tensor of order three with  $D\times C \times A $ dimensions which  contains feature fusion parameters and classifier parameters. In the other words, Eq. (1)  fuses attributes and identity features together and  then gives the output to the classifier to predict the identities. A tensor of order three looks like numbers arranged in a rectangular box and in general, an \textit{n-{th}} order tensor looks like numbers arranged in an $n$-hyper rectangle. Notation $\times$ is the multiplication of the tensor with the matrix and the left-subscript of $f(x_j,w_i)$ and $g(x_j,w_a)$ represents their mode of product. In general, the $k$-mode product of a tensor $\textbf{X}\in\mathbb{R}^{I_1\times I_2,...,\times I_N}$ with a matrix $M \in \mathbb{R}^{L \times I_k}$ is defined as follows:
\begin{equation}
 \textbf{Y}=\textbf{X}\times_k M,
\end{equation}
 where $N$ and $I_1\times I_2,...,\times I_N$ are the order and dimensions of the tensor $\textbf{X}$, respectively. The result, $\textbf{Y}$, is also a tensor with order of $N$ and dimension of ${I_1,...,I_{k-1}\times L \times  I_{k+1},...,\times I_N}$. For example, in our case, $\textbf{W}\times_1 f(x_j,w_{i})\times_3 g(x_j,w_a)$ results in a $1\times C \times 1 $  tensor, which is a row vector because the output of  $\textbf{W}\times_1 f(x_j,w_{i})$ is a $1\times C \times A $  tensor, and the product of the resulting tensor with  $ g(x_j,w_a)  $ in the third mode is a $1\times C \times 1$ tensor. Intuitively, it becomes easier if we realize that the $k$-mode product is equal to multiplying each mode-$k$ fiber of $\textbf{X}$ with the matrix $M$. Therefore, based on matrix multiplication, each element in $\textbf{Y}$ is obtained as follows:
\begin{equation}
 y_{i_1...i_{k-1}l i_{k+1}...i_N}=\sum_{i_k=1}^{I_k} x_{i_1 i_2...i_N} m_{li_k}.
\end{equation}

 \section{Limitation and Proposed Method}
 In section (3), we provided a tensor-based model for ReID which fuses identity and attribute features together. However, tensor $ \textbf{W}$ in our model includes a large number of parameters (i.e., $D\times C\times A$) which makes the training process very slow and possibly leads to overfitting as well. As a result, it might be problematic to apply $\textbf{W}$ directly in the model to predict  identities. To solve this problem, we propose two new techniques to reduce the total number of parameters which contain $\textbf{W}$ during the training.
 
The first technique is inspired by Tensor Decomposition (TD) methods \cite{kolda2009tensor}. In this technique, the original tensor $\textbf{W}$  is  approximated by  a set of matrices and one small core tensor (e.g., $\tilde{\textbf{{W}}}$) such that the number of the parameters which contain all of these components is much less than the number of the parameters which contain the original tensor. We call these matrices and small core tensor  the components of the original tensor. In the next step, the original tensor of the model (i.e., $\textbf{W}$ in Eq. (1)) is replaced by these components to predict identities with a fewer number of the parameters.   
% Here,  we call these matrices and small core tensor as the components of the original tensor
 In the second technique, we use a Structured Sparsity Learning (SSL) method \cite{huang2011learning, wen2016learning} to regularize the structure  of $\tilde{\textbf{{W}}}$ (i.e., the slices from the top, side and front modes). SSL can  effectively: (1) reduce the total number of parameters which contain  the tensor $\tilde{\textbf{{W}}}$ by zeroing all the parameters in some slices. This process leads to a compact structure from the bigger tensor $\tilde{\textbf{{W}}}$, (2)  obtain a structured sparsity from the core tensor to efficiently expedite the training process. The first  and the second techniques are described in the  section (5) and (6), respectively.

\section{Parameter Reduction Based on TD } 

Here, we explain our TD technique for  reducing the number of  parameters. In this technique, we  leverage  Higher  Order  Singular Value Decomposition (HOSVD)-based tensor decomposition algorithm \cite{bergqvist2010higher} to provide  a  theoretical  equivalent  model to the original one. Our goal is to train an equivalent  model instead of the original model, because the equivalent  model contains fewer number of the parameters than the original, which makes it feasible and stable model for training. 

In this technique, we decompose the original tensor $\textbf{W}$ to  a set of matrices and one simpler core tensor (i.e., tensor components) which can closely approximate it. The components approximate the original tensor by minimizing the reconstruction error between the original and approximated tensors. The reconstruction process is performed cyclically until our objective function converges to a close approximation of the original tensor. In reconstruction process, the values of leading left singular vectors in all three modes (i.e., top, side and front) increases gradually in each repetition. Since we do not use all the   singular vectors in each mode to reconstruct the original tensor, we significantly reduce the total number of the reconstruction parameters (i.e., all the parameters which contain the components).
\subsection{Equivalent  Model Based on TD}
Here, we describe our method based on TD which provides a  theoretical  equivalent model to the original model. The goal of parameters reduction is to reduce the total number of the parameters during training of the model to  prevent it from overfitting as well as speeding up the training process. 

In multilinear algebra, TD is a strategy that represents a tensor as a sequence of elementary operations which are performed on the other simpler tensors. Most of the TD methods are the generalized version of  matrix decomposition approaches. For example, by extending the Singular Value Decomposition (SVD) to the higher order, which is called as HOSVD, an $N$-order tensor $\textbf{X}$ can be approximated as follows:
\begin{equation}
 \tilde{\textbf{X}}=\textbf{G}\times_1 A^{(1)}\times_2 A^{(2)} \times_3...\times _N A^{(N)},
\end{equation}

 where $\tilde{\textbf{X}}$  is the approximation of the given tensor $\textbf{X}$, and  $\textbf{G}$ is a simpler tensor with much smaller size than the original tensor, $\textbf{X}$. Note that the tensor $\textbf{X}$ was decomposed by $N$ number of elementary matrices products with a much simpler tensor $\textbf{G}$. Based on  Eq. (4), the three order tensor $\textbf{W} \in \mathbb{R}^ {D \times C \times A }$ defined by our model (i.e., Eq. (1)) is decomposed as follows:
 \begin{equation}
  \tilde{\textbf{W}}=\textbf{G}\times _1 A^{(1)}\times_2 A^{(2)} \times_3 A^{(3)},
 \end{equation}  
  here,  $\textbf{G}$ is a third order tensor with $r_d \times r_c \times r_a $ dimensions; $A^{(1)}$ , $A^{(2)}$ and $A^{(3)}$  are  $r_d \times D$,   $r_c \times C$  and  $r_A \times A$ matrices, respectively.   Parameters reduction takes place by setting   $r_d$ , $r_c$ and $r_a$ to small values while minimizing the reconstruction error (i.e., $||\tilde{\textbf{W}}- \textbf{W}||$). For example, assume that $ r_d \textless \textless D $ , $ r_c \textless \textless C $  and $ r_a \textless \textless A$, then the total number of parameters reconstructing $\textbf{W}$  (i.e., $r_d \times r_c \times r_a + r_d \times D + r_c \times C + r_a \times A $)  is much smaller than the number of parameters containing $\textbf{W}$ (i.e., $D \times C \times A$).

An equivalent model is obtained by  replacing the original tensor  $\textbf{W}$ with the approximated tensor $\tilde{\textbf{W}}$ in the original model defined in Eq. (1). Here, we formulate the equivalent model  by replacing  $\textbf{W}$ with $\tilde{\textbf{W}}$ as follows: 
\begin{equation}
  \begin{aligned}
& \tilde{y_j}=\textbf{W}\times_1 f(x_j,w_i)\times_3 g(x_j,w_a) \Rightarrow{} \\
&\tilde{y_j}= \textbf{G}\times _1 A^{(1)}\times_2 A^{(2)} \times_3 A^{(3)}\times_1 f(x_j,W_i)\times_3 g(x_j,w_a).  
\end{aligned}
\end{equation}
There are two equalities in a $k$-mode product between a tensor and a matrix which we can use to simplify Eq. (6):\\

$\textit{a)\ }\textbf{X}\times _m A^{(1)}\times_n A^{(2)} =\textbf{X}\times _n A^{(2)}\times_m A^{(1)} \ (if \ m\neq n)$\\

$\textit{b)\ } \textbf{X}\times _m A^{(1)}\times_m A^{(2)} =\textbf{X}\times _m (A^{(2)} A^{(1)}) $\\ \\
Based on (a and b), Eq. (6) can be written as follows:$$ \tilde{y_j}= \textbf{G}\times _1 A^{(1)}\times_2 A^{(2)} \times_3 A^{(3)}\times_1
f(x_j,w_i)\times_3 g(x_j,w_a)\xrightarrow{a} $$ 
$$ \tilde{y_j}=\textbf{G}\times _1 A^{(1)}\times_1 f(x_j,w_i)\times_2 A^{(2)} \times_3 A^{(3)}\times_3 g(x_j,w_a)\xrightarrow{b}$$
$$ \tilde{y_j}=\textbf{G}\times _1 (A^{(1)} _1 f(x_j,w_i))\times_2 A^{(2)} \times_3 (A^{(3)} _3 g(x_j,w_a)).$$
Based on the following theorem in the tensor decomposition, tensor multiplications can be simplified as follows:
\begin{equation}
\begin{aligned}
&\textbf{X}=\textbf{G}\times_1 A^{(1)}\times_2 A^{(2)} \times_3...\times _N A^{(N)} \Leftrightarrow \\
&\textbf{X}_k= A^{(k)}\textbf{G}_k(A^{(N)}\otimes ...\otimes A^{(k-1)} \otimes A^{(k+1)}...\otimes A^{(1)})^\top,
\end{aligned}
\end{equation}
\setlength{\parindent}{0pt} where $\otimes$ denotes the Kronecker product; $\textbf{X}_k$ is the mode-$k$ unfolding (mode-$k$ matricization) of the tensor $\textbf{X}$. The mode-k unfolding organizes the mode-$k$ fibers of $\textbf{X}$ as columns into a matrix. A fiber is a generalization of columns to tensors.

\textbf{The Kronecker product} is a generalization of outer product for matrices.  It is an operation on two matrices with  arbitrary sizes which results in a block matrix. The Kronecker product is mathematically formed by the direct product of two matrices and there are no learnable parameters in this operation. Thus, the chance of overfitting for doing this operation is very negligible. Assume that $A$ is an $m \times n$ matrix and $B$ is a $p \times q$ matrix, then the Kronecker product of $A$ and $B$ is $mp \times nq$ block matrix:$$ A\otimes B=\begin{bmatrix}
    a_{11}B       & \dots & a_{1n}B\\
    \vdots       & \vdots & \vdots \\
    a_{m1}B       & \dots & a_{mn}B
\end{bmatrix}. $$ By considering Eq. (7), Eq. (1) can be written as follows:
\begin{equation}
{\tilde{y_j}}=\underbrace{A^{(2)} \textbf{G}_2}_\text{Classifier} \underbrace{ (A^{(3)} g(x_j,w_a)\otimes A^{(1)}f(x_j,w_i))^\top.}_\text{Fused Feature}
\end{equation}
Eq. (8) is the equivalent model to the original model in Eq. (1) which has identical format of the model. This equation indicates that the approximated tensor $\tilde{\textbf{W}}$ similar to the original tensor $\textbf{W}$ defined in Eq. (1) performs feature fusion as well as identity classification.

 \begin{figure}[t]
\centering
{\includegraphics [scale=.8]{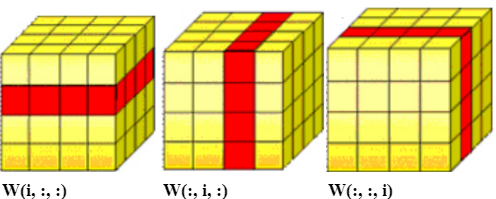}}
\caption{SSL on tensor: Grouping from top, side, and front.}\label{fig_sparse} 
\end{figure}
\subsection{Components Estimation from Original Tensor}

In section 5.1, we provided a TD technique to establish a theoretical equivalent model to the original model. In this section, we explain an algorithm which is used to estimate the decomposed components including $A^{(1)}, A^{(2)}, A^{(3)} \text{and } \textbf{G} $. Here, we aim to approximate the original tensor $ \textbf{W} $ with the minimal reconstruction error, while at the same time,  reducing the total number of parameters which contain the decomposed components. In this method, we start with  the HOSVD algorithm which is considered as a  matrix SVD generalization method. Since the matrices $A^{(k)}$ are orthogonal and tensor $\textbf{G}$ is “ordered” and "all-orthogonal",  HOSVD returns the decomposed components $A^{(1)}, A^{(2)}, A^{(3)}$ and $\textbf{G} $ as described in Algorithm 1.

\begin{algorithm}[t]
\caption{HOSVD}
\begin{algorithmic}[1]
\For{\text{k=1, 2, 3}}
  \State{$A^{(k)}\Leftarrow \text{left orthogonal matrix of the SVD from\  } \textbf{W}_k $}
\EndFor\\
$\textbf{G}\xleftarrow{} \textbf{W}\times_1 (A^{(1)})^\top \times_2 (A^{(2)})^\top \times_3 (A^{(3)})^\top $\\
\textbf{return:}$\ \textbf{G}$, $A^{(1)}$, $A^{(2)}$, $A^{(3)}$
\end{algorithmic}
\end{algorithm}  
In Algorithm 1, $\textbf{W}_k$ denotes the unfolded tensor $\textbf{W}$ in the mode $k$. Tensor $\textbf{W}$ can be approximated by truncating the matrices $ A^{(k)} $. The truncated HOSVD measured by the norm of the difference (i.e., $||\textbf{W}-\textbf{G}\times_1 A^{(1)}\times_2 A^{(2)}\times_3 A^{(3)} || $) does not provide the best fit. However,  the truncated HOSVD is a good starting point for the other TD algorithms \cite{symeonidis2016matrix, bergqvist2010higher}. The higher order orthogonal iteration (HOOI) algorithm finds the optimal approximation $\tilde{\textbf{W}}$ (with respect to the Frobenius norm loss) by iterating the alternating truncation and SVD until convergence \cite{symeonidis2016matrix, bergqvist2010higher}. If  $A^{(k)}$ is truncated to have $r_k$ columns, then the HOOI solution can be obtained by  Algorithm 2 \cite{symeonidis2016matrix, bergqvist2010higher}.
\begin{algorithm}[t]
\text{\textbf{initialize}} $\ \textbf{G}$, $A^{(1)}$, $A^{(2)}$, $A^{(3)}$\text{\ by using \textbf{Algorithm 1}}
\caption{HOOI}
\begin{algorithmic}[1]
\While{not converged}
\For{\text{k=1, 2, 3}}
  \State{$\textbf{Y}\xleftarrow{} \textbf{W}\times_{1} (A^{(1)})^\top \times_{2} (A^{(2)})^\top \times_{3} (A^{(3)})^\top$}
  \State{$A^{(k)}\Leftarrow \text{$r_k$ leading left singular vectors of\  } \textbf{Y}_k $}
\EndFor\\
$\textbf{G}\xleftarrow{} \textbf{W}\times_1 (A^{(1)})^\top \times_2 (A^{(2)})^\top \times_3 (A^{(3)})^\top $
\EndWhile\\
\textbf{return:}$\ \textbf{G}$, $A^{(1)}$, $A^{(2)}$, $A^{(3)}$
\end{algorithmic}
\end{algorithm}

\section{Parameter Reduction Based on SSL}
The regularization based on SSL is a class of methods and an area of research in statistical learning theory that extends and generalizes sparsity regularization learning methods. Both sparsity and SSL aim to exploit the assumption that the output variable (i.e., response, or dependent variable) can be represented by a reduced number of variables in the input space (i.e., the domain, space of features or explanatory variables). SSL methods focus on selecting the input variables that best describe the output. These methods generalize and extend sparsity regularization methods by allowing for optimal selection over structures like groups or networks of input variables.

Here, we introduce the second technique to reduce  parameters of the tensor. This technique is based on SSL \cite{huang2011learning,wen2016learning} which is applied on the slices of the tensor in each mode (Fig. \ref{fig_sparse}, slices of the tensor in front, top and side modes)  to regularize the structure of the tensor. In our case , SSL adds structural sparsity regularization terms on the estimated tensor $\tilde{\textbf{W}}$ for each of the three modes into our loss function  to learn a compact structure.  SSL can effectively: (1) reduce the total number of parameters which contain the  approximated tensor by zeroing all the parameters in some slices, and (2) obtain a structured sparsity from the estimated tensor to efficiently expedite the training process. In the following section, we will introduce our loss function which uses the SSL method to regularize the estimated tensor. 

\section{Sparse Tensor-Based Model for ReID}
%The contrastive loss function is  widely used in the literature of  deep neural network tasks such as invariant feature learning
 Here, we provide a loss function for person ReID  which considers SSL on the parameters of the tensor. The total loss function is formulated in Eq. (9). We use a multi-task loss function where $\mathcal{L}_c$ (terms 1 and 4) is the soft-max cross entropy loss which is used for prediction tasks (attribute and identity prediction) and $\mathcal{L}_{con}$ (term 2) is the contrastive loss which is used to generate discriminative fused features for ReID. The goal of contrastive loss is to bring genuine pairs close to each other in the feature space while pushing them away if they are imposter pairs. Variable $\tilde{y_j}$ is the  label corresponding to samples $j$  predicted by our model, and ${y_j}$ is the ground truth label. Values $n$ and $s$ are the number of samples in each training batch and number of attributes, respectively. Variable $\tilde{l_{t,j}}$  is the predicted  attribute by the  network corresponding to the \textit{t-th} attribute and the \textit{j-th} training sample, respectively and ${l_{t,j}}$ is its ground truth label. Parameters $\lambda_1$, $\lambda_2$, $\lambda_3$ are the balancing parameters between the different losses in the total loss function.
 % Since our model is an end to end framework, we predict attributes and identity of the people once at the time. Therefore, our total loss function includes attribute and identity prediction losses together and updates  parameters of the predictors (attribute and identity) trough the training of the model.

 \begin{equation}
 \begin{split}
    & \mathcal{L}(\tilde{\textbf{W}}, w_{i},w_a)=...\\ & \underbrace{\sum_{j=1}^n \mathcal{L}_c(A^{(2)},\textbf{G}, w_{i},\tilde{y_j},y_j)}_\text{(1)}+\lambda_1 \underbrace{ \sum_{j,k=1}^n \mathcal{L}_{con}(A^{(1)},A^{(3)}, {y_j},{y_k}) }_\text{(2)}\\ & + \lambda_2 \underbrace{(\sum_{p=1}^D ||\tilde{\textbf{W}}_{p,:,:}||_2  +  \sum_{q=1}^C ||\tilde{\textbf{W}}_{:,q,:}||_2 +  \sum_{s=1}^A ||\tilde{\textbf{W}}_{:,:,s}||_2 )}_\text{(3)}...\\ &+\lambda_3\underbrace{\sum_{t=1}^{s}\sum_{j=1}^n \mathcal{L}_c(w_{a},\tilde{l_{tj},l_{tj})}}_\text{(4)}.
    \end{split}
 \end{equation}

%This loss function  updates  $A^{(2)}$  and $f$  network  parameters  during training; however, it is not used to update attribute predictor parameters ($ w_a$) because in our model, attributes are predicted by the $g$ network with a different cost function. In the other words, $\tilde{\textbf{W}}$ in () fuses attribute and identity features and use it for ReID without influencing attribute predictor parameters. It is worth of mentioning that  we can extend our model to multi-task learning problem to train attribute prediction and ReID tasks jointly by sharing $\tilde{\textbf{W}}$ parameters which will be our feature work.
 As discussed in Section 5, $\textbf{W}$ is decomposed to components in a way that some components (i.e, $A^{(1)}$ and $A^{(3)}$, see Eq. (8)) fuse two set of features while the other components (i.e., $A^{(2)}$ and $\textbf{G}$ in Eq. (8)) perform as an identifier. Here, term (1)  is the  classification loss which is used to train identity network, and identifier components of the tensor $\tilde{\textbf{W}}$ (i.e., $A^{(2)}$ and $\textbf{G}$). Suppose that the model prediction for the \textit{j-th} sample is  $\tilde{y_j} $, then the softmax-cross entropy loss, $\mathcal{L}_c$ for sample $j$ is obtained as: 
 \begin{equation}
  \mathcal{L}_c(A^{(2)}, w_{i},w_a)=\sum_{i=1}^C y_j^{(i)} \log(\tilde{y_j}^{(i)}),
\end{equation}
  where  $y_j^{(i)}$ and $\tilde{y_j}^{(i)}$ indicate the \textit{i-th}  elements of the label vectors $y_j$ and $ \tilde{y_j} $, respectively. 
Term (2) is a contrastive loss function which forces $A^{(1)}$ and  $A^{(3)}$ (i.e., components of the tensor in Eq. (8)) to bring  genuine pairs close to each other in the fused features space while pushing away the imposter pairs from each other.  The value of $\mathcal{L}_{con}$ for sample $j$ and $k$ is obtained as follows: 
\begin{equation}
\begin{split}
&\mathcal{L}_{con}\tilde{(\textbf{W}}, w_{i}, {y_j},{y_k})= \\ & 
  (1-Y)\frac{1}{2}(d)^2 + (Y)\frac{1}{2}(\mbox{max}(0,m-d))^2,
  \end{split}\label{eq:5}
\end{equation}

where $d$ is the distance between two fused features for samples $j$ and $k$. $Y$ is equal to one if two samples are genuine and zero if samples are imposter. Variable $m$ is a margin value which is greater than zero. A margin indicates that imposter pairs that are beyond this margin will not contribute to the loss.

 Terms (3) is the sparsification term for regularizing  the tensor $\tilde{\textbf{W}}$. This term is the structured sparsity regularization term. Since Group Lasso can effectively make  all the weights in some groups zero \cite{yuan2006model}, we use it in our SSL method for regularizing $\tilde{\textbf{W}}$. The regularization of group Lasso on a set of weights in top , front and side  modes are $ \sum_{p=1}^{r_d} ||\tilde{\textbf{W}}_{p,:,:}||_2 $ , $\sum_{s=1}^{r_a} ||\tilde{\textbf{W}}_{:,:,s}||_2$ and $\sum_{q=1}^{r_c} ||\tilde{\textbf{W}}_{:,q,:}||_2$, respectively where $r_d$, $r_c$ and $r_a$ are the number of tensor slices in top, side and front, respectively. Each weight $\tilde{\textbf{W}}_{ijk}$   in $\tilde{\textbf{W}}$ is indexed by three indices $i,j$ and $k$ and $\tilde{\textbf{W}}_{i,:,:}$ represents all the weights which have the first index of $i$.  This set of weights constructs the \textit{i-th} slice of the tensor from top. For example, the term $||\tilde{\textbf{W}}_{i,:,:}||_2$ is obtained as follows:
 \begin{equation}
     ||\tilde{\textbf{W}}_{i,:,:}||_2=\sqrt{\sum_{j=1}^{r_c}\sum_{k=1}^{r_a} \tilde{\textbf{W}}_{ijk}^2}.  
 \end{equation}

\section{Experiments and Discussion}
In this section, we initially describe our CNN architecture, training setup and the benchmarks that we have used in our experiments. Here, we experimentally evaluate our model with the following guidelines:  1) We provide an analysis on the hyperparameter tuning, and sensitivity of our model performance to the hyperparameters of the model. 2) We report accuracy of the attribute prediction via our backbone CNN model. 3) We compare our model with the baselines and other state-of-the-art methods which are both attribute and non-attribute based ReID approaches. 4) We study the contribution of the SSL regularization and TD techniques in our tensor-based model for ReID. 5)  Many attributes features generally focus  on describing the local properties of the identities in the images and then the features obtained from the attributes potentially provide complementary information along with the identities features which represent the global properties of a person. In this experiment, we study the attention of the attributes features on the identity features in our tensor-based model for ReID. 6) We study effectiveness of the SSL regularization in our model on the speedup and also, we report the percentage of the sparsity for reducing the parameters of the tensor. 7) We  qualitatively investigate the feature representation of  our  tensor-based  model  for  ReID.  8) We investigate the level of our model confidence on the true positive pairs in  ReID. Here, the goal is to study whether  or not reducing the number of the parameters in our tensor-based model has a positive impact on the confidence of the model for the ReID. 9) We discuss the limitation of our model, and then provide a solution to address it. 
\begin{figure*}[t]
\centering
\includegraphics[scale=0.36]{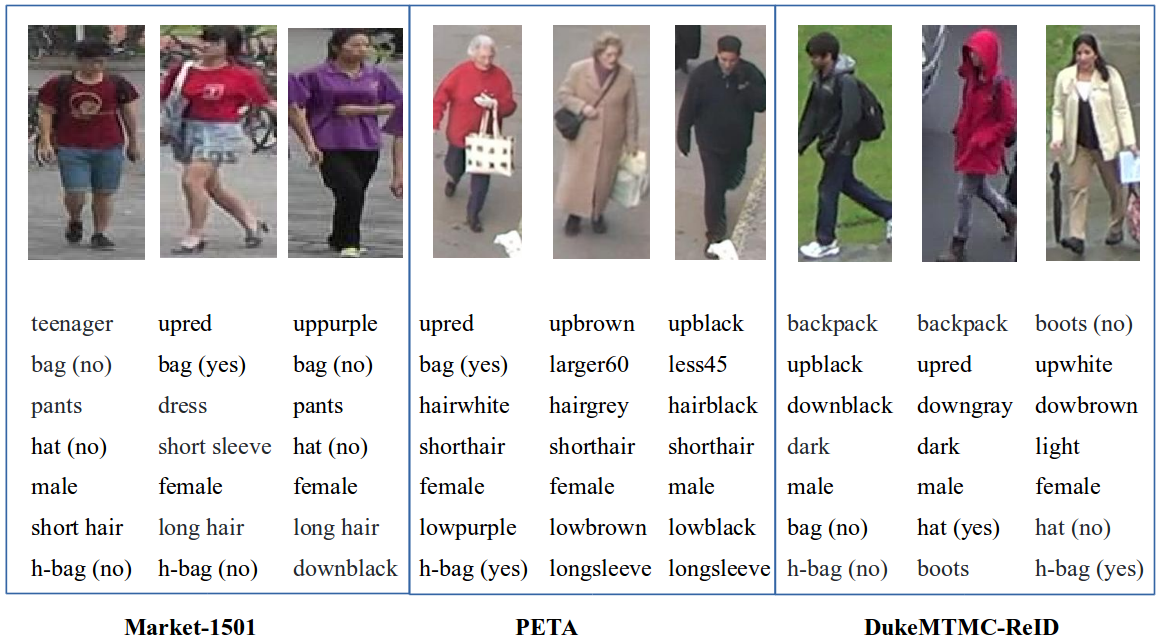}
\caption{Examples  of  images from Market-1501, PETA, and DukeMTMC-ReID datasets annotated by attributes.}\label{sample}
%\vspace{-5.5mm}
\end{figure*}
\begin{table*}
\centering
\small
\caption{Rank 1 accuracy for hyperparameters tuning on the Market-1501, DukeMTMC-ReID and PETA datasets.}
\scalebox{0.82}{\begin{tabular}{|c|c|c|c|c|c|c|c|c|c|c|c|c|c|}
 \hline
\multicolumn{1}{|c}{\multirow{1}{*}{}} &\multicolumn{4}{|c}{\multirow{1}{*}{Contrastive}} &\multicolumn{4}{|c|}{SSL}&\multicolumn{4}{|c|}{Attribute prediction}\\ \cline{2-13}  
\multicolumn{1}{|c}{\multirow{1}{*}{Hyperparameter}} &\multicolumn{1}{|c}{\multirow{1}{*}{$\lambda_1=0.01$}}&\multicolumn{1}{|c}{\multirow{1}{*}{$\lambda_1=0.1$}}&\multicolumn{1}{|c}{\multirow{1}{*}{$\lambda_1=1$}}&\multicolumn{1}{|c}{\multirow{1}{*}{$\lambda_1=10$}} &\multicolumn{1}{|c}{\multirow{1}{*}{$\lambda_2=0.01$}}&\multicolumn{1}{|c}{\multirow{1}{*}{$\lambda_2=0.1$}}&\multicolumn{1}{|c|}{\multirow{1}{*}{$\lambda_2=1$}}&\multicolumn{1}{|c}{\multirow{1}{*}{$\lambda_2=10$}}&\multicolumn{1}{|c}{\multirow{1}{*}{$\lambda_3=0.01$}}&\multicolumn{1}{|c}{\multirow{1}{*}{$\lambda_3=0.1$}}&\multicolumn{1}{|c|}{\multirow{1}{*}{$\lambda_3=1$}}&\multicolumn{1}{|c|}{\multirow{1}{*}{$\lambda_3=10$}}\\ \hline \hline
Market-1501 & $\textbf{96.46\%}$& $95.09\%$ &$95.42\%$ &$ 94.67\%$& $95.15\%$& $96.18\%$& $\textbf{96.69\%}$& $95.84\%$& $94.75\%$& $95.73\%$& $\textbf{96.01\%}$& $95.9\%$\\ \hline
DukeMTMC-ReID& $93.12 \%$ &$93.99\%$ &$94.53 \%$ &$\textbf{95.76 \% } $&$92.52 \%$&$94.74\%$&$94.62 \%$ &$\textbf{95.34 \%}$ &$\textbf{95.56 \%}$& $94.97 \%$ &$94.37 \% $&$95.29\%$\\ \hline
PETA& $80.34 \%$ &$\textbf{81.06 \%}$ &$79.96\%$ &$80.12\%$&$79.88 \%$&$80.75\%$&$\textbf{81.34 \%}$ &$81.25 \%$ &$\textbf{81.48 \%}$& $80.89  \%$ &$81.18 \% $&$80.2 \%$\\ \hline
\end{tabular}}
\label{table_t1}
\end{table*}
\subsection{CNN Architecture} 

We use a ResNet-50 \cite{he2016deep} as our CNN backbone for both the attribute and identity networks in Fig. \ref{fig1}. We use batch normalization after each convolutional layer, and before performing ReLU activation function. Batch normalization potentially helps to obtain faster learning. Moreover, batch normalization allows us to use a higher learning rate, which potentially results in another boost in speed. We use  an Adam optimizer  with default hyper-parameters values ($\epsilon=10^{-3}$, $\beta_1 = 0.9$, $\beta_2 = 0.999$) to train  our model. The batch size in all experiments is fixed to 256 and the framework is implemented in PyTorch. We performed our experiments on two GeForce GTX TITAN X 12GB GPU.

\subsection{Training the Framework}
In Sections (5) and (6), we used SSL and TD approaches to create an accurate and stable learning problem for our original model in Fig. \ref{fig1}.  To train the parameters of the attributes and identity networks, a large amount of images is needed. Thus, we initialize the parameters of these networks by pre-training on the ImageNet dataset. Note that the compared methods  are also pre-trained on the ImageNet dataset for a fair comparison. We initialize  the parameters of the tensor using a uniform distribution. Since our model is an end to end framework, all the parameters of the two networks in Fig. \ref{fig1} are updated simultaneously in every training step. For each training batch, we minimize term (4) in Eq. (9) (i.e., attribute predictor) with respect to the  parameters of the attribute network and term (1) with respect to the parameters of the identity network and components of the tensor which performs the prediction task. We optimize term (2) with respect to the parameters of the  components which fuse the identity and attribute features (i.e., $A^{(1)}, A^{(3)}$) to enforce these parameters to generate more discriminative features. Other than these terms, a SSL constraint on the tensor parameters is also applied in the loss function to regularize the structure of the tensor. We iterate this training procedure until there is no improvement of the identification performance on the training set.
\begin{table*}[t]
    \centering
    \caption{Accuracy of the Attribute prediction on (a) Market-1501 , and (b) DukeMTMC-ReID }
\begin{tabular}{ ccc }   % top level tables, with 2 columns

(a) Market-1501 & (b) DukeMTMC-ReID 1 \\  

% leftmost table of the top level table
\begin{tabular} {|l*{18}{|c}r}

\hline
\centering
\rotatebox{90} {Attributes} &
\rotatebox{90}{S.clth}& 
\rotatebox{90}{B.pack} & 
\rotatebox{90}{L.low} & 
\rotatebox{90}{L.slv} &
\rotatebox{90}{H.bag} & 
\rotatebox{90}{Bag} &
\rotatebox{90}{Hat} &
\rotatebox{90}{C.up} &
\rotatebox{90}{C.low} &
\rotatebox{90}{Gender} &
\rotatebox{90}{Hair} &
\rotatebox{90}{Age} 
 \\
\hline

\rotatebox{90} {Acc} & \rotatebox{90}{$93.23\%$} & \rotatebox{90}{$84.15\%$} & \rotatebox{90}{$90.95\%$} & \rotatebox{90}{$89.91\%$} & \rotatebox{90}{$87.23\%$} & \rotatebox{90}{$77.58\%$} & \rotatebox{90}{$92.16\%$} & \rotatebox{90}{$73.14\%$} & \rotatebox{90}{$72.16\%$} & \rotatebox{90}{$86.95\%$} & \rotatebox{90}{$81.9\%$} & \rotatebox{90}{$84.04\%$} \\
   \hline
\end{tabular} &  % starting rightmost sub table
% table 2
\begin{tabular} {|l*{11}{|c}r}
\hline
\centering
\rotatebox{90} {Attributes} &
\rotatebox{90}{B.pack}& 
\rotatebox{90}{H.bag} & 
\rotatebox{90}{L.slv} &
\rotatebox{90}{Bag} & 
\rotatebox{90}{Boots} &
\rotatebox{90}{Gender} &
\rotatebox{90}{Hat} &
\rotatebox{90}{C.shoes} &
\rotatebox{90}{C.up} &
\rotatebox{90}{C.low} 
 \\
\hline

 \rotatebox{90}{Acc} &  \rotatebox{90}{$78.25\%$} & \rotatebox{90}{$90.86\%$} & \rotatebox{90}{$88.11\%$} & \rotatebox{90}{$80.92\%$} & \rotatebox{90}{$88.56\%$} & \rotatebox{90}{$80.12\%$} & \rotatebox{90}{$82.1\%$} & \rotatebox{90}{$83.24\%$} & \rotatebox{90}{$75.56\%$} & \rotatebox{90}{$70.05\%$} \\
 \hline
\end{tabular} 

\end{tabular}

\label{tab_2}

\end{table*}

% \begin{figure*}[t]
% \centering
%  \subfigure[\label{fig3a}]{\includegraphics [scale=.36]{Figure_3g.png}}
%  \subfigure[\label{fig3b}]{\includegraphics [scale=.4]{Figure_1g.png}}
%  \subfigure[\label{fig3a}]{\includegraphics [scale=.4]{Figure_2g.png}}

% \caption{Attribute prediction accuracy: a) Market-1501,  b) DukeMTMC-ReID datasets.}\label{fig2} 
% \end{figure*}
\begin{table}
\caption{Accuracy of the attribute prediction on PETA dataset.}
\begin{tabular} {|l*{27}{|c}r}

\hline

\rotatebox{90} {Attributes} &
\rotatebox{90}{Leather-Shoes}& 
\rotatebox{90}{Logo} & 
\rotatebox{90}{Long-hair} &
\rotatebox{90}{Male} & 
\rotatebox{90}{Messenger-Bag} &
\rotatebox{90}{Sandals} &
\rotatebox{90}{Shoes} &
\rotatebox{90}{Shorts} &
\rotatebox{90}{Short-Sleeve} &
\rotatebox{90}{Skirt} &
\rotatebox{90}{Sneaker} &
\rotatebox{90}{Stripes} 
 \\
\hline
 \rotatebox{90}{Acc} & 
 \rotatebox{90}{$58.9\%$ } & \rotatebox{90}{$96.54\%$} & \rotatebox{90}{$72.46\%$} & \rotatebox{90}{$40.3\%$} & \rotatebox{90}{$60.54\%$} & \rotatebox{90}{$96.14\%$} & \rotatebox{90}{$61.4\%$} & \rotatebox{90}{$97.74\%$} & \rotatebox{90}{$91.45\%$} & \rotatebox{90}{$95.44\%$} & \rotatebox{90}{$82.9\%$} & \rotatebox{90}{$97.94\%$}\\
 \hline

 \multicolumn{10}{c}{}\\
 
\end{tabular}

\scalebox{0.99}{\begin{tabular}{|l*{18}{|c}}
\hline
\rotatebox{90}{Attribute}&
\rotatebox{90}{Sunglasses}&
\rotatebox{90}{Trousers}&
\rotatebox{90}{Tshirt} &
\rotatebox{90}{UpperOther} &
\rotatebox{90}{V-Neck} &
\rotatebox{90}{Age16-30} &
\rotatebox{90}{Age31-45} &
\rotatebox{90}{Age46-60} &
\rotatebox{90}{AgeAbove61} &
\rotatebox{90}{Backpack}&
\rotatebox{90}{CarryingOther}&
\rotatebox{90}{Formal-lower} \\
\hline
\rotatebox{90}{Acc}&\rotatebox{90}{$92.85\%$} & \rotatebox{90}{$52.96\%$} & \rotatebox{90}{$92.4\%$} & \rotatebox{90}{$25.76\%$} & \rotatebox{90}{$97.92\%$}& \rotatebox{90}{$59.73\%$} & \rotatebox{90}{$64.95\%$} & \rotatebox{90}{$82.84\%$} & \rotatebox{90}{$87.48\%$} & \rotatebox{90}{$ 82.86\%$} & \rotatebox{90}{$79.64\%$}&\rotatebox{90}{$87.45\%$}\\
 \hline
 \multicolumn{10}{c}{}\\
\end{tabular}}

\scalebox{0.99}{\begin{tabular} {|l*{18}{|c}}

\hline
\centering
\rotatebox{90} {Attributes} &
\rotatebox{90}{Formal-upper} &
\rotatebox{90}{Hat} &
\rotatebox{90}{Jacket} &
\rotatebox{90}{Jeans} &
\rotatebox{90}{Muffler} &
\rotatebox{90}{No-accessory} &
\rotatebox{90}{No-carrying} &
\rotatebox{90}{Plaid} & 
\rotatebox{90}{PlasticBags}&
\rotatebox{90}{CarryingOther}& 
\rotatebox{90}{Casual-lower} & 
\rotatebox{90}{Casual-upper}
 \\
\hline
 \rotatebox{90}{Acc} & \rotatebox{90}{$86.97\%$} & \rotatebox{90}{$82.18\%$} & \rotatebox{90}{$92.53\%$} & \rotatebox{90}{$66.54\%$} & \rotatebox{90}{$82.91\%$} & \rotatebox{90}{$57.66\%$} & \rotatebox{90}{$76.83\%$} & \rotatebox{90}{$97.17\%$} & \rotatebox{90}{$86.94\%$}&\rotatebox{90}{$79.64\%$} & \rotatebox{90}{$87.51\%$} & \rotatebox{90}{$89.25\%$}\\
 \hline
\end{tabular}}
\label{tab_3}
\end{table}

\textbf{Evaluation Protocols:}  we use Cumulative Matching Characteristics (CMC) at rank1 and mean Average Precision (mAP) as our evaluation metrics to compare the performance of our proposed method with the baselines and other state-of-the-art methods.

\subsection{Benchmarks}
\textbf{Market-1501} dataset \cite{zheng2015scalable} includes 19.7k images of 751
identities for training and 13.3k images of 750 identities for
testing (3368 query images and 16364 gallery images). Each image in this dataset is annotated by 27 attributes \cite{lin2019improving}. 

\textbf{DukeMTMC-ReID}  is a subset of DukeMTMC  \cite{zheng2017unlabeled} containing 16.5k training images of 702 identities and 19.9k test images of 702
identities (2228 query images and 17661 gallery images). Each image in this dataset is annotated by 23 attributes \cite{lin2019improving}.

\textbf{PETA}  is annotated by 61  binary attributes and 4 multi-class attributes for 19k images. Here, we follow \cite{deng2014pedestrian,lin2019improving}, and we use the 35 most important  attributes for person ReID in our experiments. PETA contains few samples for some identities, and in some cases, there is only one sample for some identities. Thus, following previous work \cite{lin2019improving}, we re-split this dataset to 17.1k
images of 4,981 identities (9.5k images of 4,558 identities for training, 423 images for query, and 7.2k images for gallery) to evaluate our model. Examples of these three datsets have been illustrated in Fig. \ref{sample}.
\subsection{Hyperparameters Tuning}
Our model contains three hyperparameters $\{\lambda_1, \lambda_2, \lambda_3\}$ which  control the role of the contrastive loss, SSL, and attribute prediction objectives in the total loss function in Eq. (9), respectively. For each of these hyperparameters, we choose the values to be $\{0.01, 0.1, 1, 10\}$. Here, we use $10\%$ of the training data for each dataset as our validation set, and after we find the best values for each hyperparameter, we revert the validation set to the training set and then train the model again for the testing. Our experimental results based on CMC  at  rank1   in Table. \ref{table_t1} indicate that the best values for  $\lambda_1$,  $\lambda_2$, and  $\lambda_3$ on the Market-1501 dataset are $0.01$, $0.1$, and $1$, on the DukeMTMC-ReID dataset are $10$, $10$, and $0.01$, and on the PETA dataset are $0.1$, $1$, and $0.01$, respectively. Note that in all of our experiments, we set the margin in the contrastive loss to 1. Moreover, it is worth mentioning that the results in Table \ref{table_t1} indicate that our model is not significantly sensitive to the hyperparameters of our model.

\begin{table}[t]
\centering
\caption{Comparison with ReID methods on the Market1501.}
\scalebox{0.98}{\begin{tabular}{c| c c c}
 \hline
\multicolumn{1}{c|}{Method} &\multicolumn{1}{c}{Backbone} &\multicolumn{1}{c}{mAP}&\multicolumn{1}{c}{rank 1}\\ [0.5ex] 
 \hline \hline
\textbf{Our method}  & ResNet50&$\textbf{90.16}$&$\textbf{98.38}$\\
Our method w/o SSL & ResNet50&$88.47$&$96.95$\\
Our method w/o TD  & ResNet50&$86.13$&$95.89$\\
Our method w/o (TD+SSL)  & ResNet50&$85.94$&$93.94$\\
Baseline (1)  & ResNet50&$59.9$&$79.24$\\
Baseline (2) & ResNet50&$24.48$&$50.45$\\
\hline
SCSN (4 stages) \cite{chen2020salience} & ResNet50&$88.30$&$92.40$\\
SCSN (3 stages) \cite{chen2020salience} & ResNet50 &$88.50$&$95.70$\\
ABDNet \cite{chen2019abd} & ResNet50 &$88.28$&$95.60$\\
 Pyramid \cite{zheng2019pyramidal} & ResNet101 &$88.20$&$95.70$\\
 DCDS \cite{alemu2019deep} & ResNet101 &$85.80$&$94.81$\\
 APR (w/o attri) \cite{lin2019improving} & ResNet50 &$58.74$&$81.03$\\
 APR (w/o ARM) \cite{lin2019improving} & ResNet50 &$66.59$&$85.71$\\
 APR \cite{lin2019improving} & ResNet50 &$66.89$&$87.04$\\
MHN (PCB) \cite{chen2019mixed} & ResNet50 &$85.00$&$95.10$\\
 BFE \cite{dai2019batch}& ResNet50 &$86.20$&$95.30$\\
 CASN (PCB) \cite{zheng2019re} & ResNet50 &$82.80$&$94.40$\\
 AANet \cite{tay2019aanet} & ResNet152 &$83.41$&$93.93$\\
 IANet \cite{hou2019interaction} & ResNet50 &$83.10$&$94.40$\\
VPM \cite{sun2019perceive} & ResNet50 &$80.80$&$93.00$\\
PSE+ECN \cite{sarfraz2018pose}& ResNet50 &$80.50$&$90.40$\\
PCB+RPP \cite{sun2018beyond} & ResNet50 &$81.60$&$93.80$\\
PCB \cite{sun2018beyond}& ResNet50 &$77.40$&$92.30$\\
DuATM \cite{si2018dual} & DenseNet121 &$76.60$&$91.40$\\
Pose-Transfer \cite{liu2018pose} & DenseNet169 &$56.90$&$78.50$\\
SPReID \cite{kalayeh2018human} & ResNet152 &$83.36$&$93.68$\\
Tricks \cite{luo2019bag} & SEResNet101 &$87.30$&$94.60$\\
Mancs \cite{wang2018mancs} & ResNet50 &$82.30$&$93.10$\\
PAN \cite{zheng2018pedestrian}& ResNet50 &$63.40$&$82.80$\\
SVDNet \cite{sun2017svdnet} & ResNet50 &$62.10$&$82.30$
%\vspace{-12mm}
\end{tabular}}
\label{tab_4}
\end{table}

\begin{table}[t]
\centering
\caption{Comparison with ReID methods on DukeMTMC-ReID.}
\scalebox{1}{\begin{tabular}{c| c c c}
 \hline
\multicolumn{1}{c|}{Method} &\multicolumn{1}{c}{Backbone} &\multicolumn{1}{c}{mAP}&\multicolumn{1}{c}{rank 1}\\ [0.5ex] 
 \hline \hline
\textbf{Our method}  & ResNet50&$\textbf{84.74}$&$\textbf{95.42}$\\
Our method w/o SSL  & ResNet50&$82.11$&$93.52$\\
Our method w/o TD  & ResNet50&$79.93$&$90.95$\\
Our method w/o (TD+SSL)  & ResNet50&$78.24$&$89.36$\\
Baseline (1)  & ResNet50&$45.14$&$67.88$\\
Baseline (2) & ResNet50&$26.36$&$50.45$\\

\hline
SCSN (4 stages) \cite{chen2020salience} & ResNet50&$79.00$&$91.00$\\
SCSN (3 stages) \cite{chen2020salience} & ResNet50 &$79.00$&$90.10$\\
Pyramid \cite{zheng2019pyramidal} & ResNet101 &$79.00$&$89.00$\\
ABDNet \cite{chen2019abd} & ResNet50 &$78.60$&$89.00$\\
APR (w/o ARM) \cite{lin2019improving} & ResNet50 &$54.79$&$73.56$\\
APR \cite{lin2019improving} & ResNet50 &$55.56$&$73.92$\\
MHN (PCB) \cite{chen2019mixed} & ResNet50 &$77.20$&$89.10$\\
BFE \cite{dai2019batch}& ResNet50 &$75.90$&$88.90$\\
CASN (PCB) \cite{zheng2019re} & ResNet50 &$73.70$&$87.70$\\
DCDS \cite{alemu2019deep} & ResNet101 &$75.50$&$87.50$\\
AANet \cite{tay2019aanet} & ResNet152 &$74.29$&$87.65$\\
PSE+ECN \cite{sarfraz2018pose}& ResNet50 &$75.70$&$84.50$\\
IANet \cite{hou2019interaction} & ResNet50 &$73.40$&$83.10$\\
VPM \cite{sun2019perceive} & ResNet50 &$72.60$&$83.60$\\
DuATM \cite{si2018dual} & DenseNet121 &$64.60$&$81.80$\\
PCB+RPP \cite{sun2018beyond} & ResNet50 &$69.20$&$83.30$\\
SPReID \cite{kalayeh2018human} & ResNet152 &$73.34$&$85.95$\\
Pose-Transfer \cite{liu2018pose} & DenseNet169 &$56.90$&$78.50$\\
Tricks \cite{luo2019bag} & SEResNet101 &$78.00$&$87.50$\\
Mancs \cite{wang2018mancs} & ResNet50 &$82.30$&$93.10$\\
SVDNet \cite{sun2017svdnet} & ResNet50 &$56.80$&$76.70$\\
PAN \cite{zheng2018pedestrian}& ResNet50 &$51.51$&$71.59$
%\vspace{-12mm}
\end{tabular}}
\label{tab_5}
\end{table}
\subsection{Results on Attribute Prediction}
Here, we report the accuracy of the attribute prediction on the Market-1501 and DukeMTMC-ReID, PETA datasets using our attribute network.  We use the same training and testing split used for the person ReID in this study. We have been naive on the attribute prediction task in this work, and simply used the ResNet50 pre-trained on the ImageNet. Table \ref{tab_2} (a) \& Table \ref{tab_2} (b)  indicate the attribute prediction accuracy on the Market-1501 and  DukeMTMC-ReID datasets, respectively.  Here, “S.clth”, “B.pack”, “L.low”, “L.slv”, “H.bag”, “C.up”,
“C.low” “C.shoes” stand for style of clothing, backpack, length of lower-body clothing, length of sleeve,  handbag, color of upper-body clothing and color of lower-body clothing, color of shoes, respectively. Table  \ref{tab_3} indicates attribute prediction accuracy on the PETA dataset. Note that we have been
naive in attribute prediction, which implies that ReID performance potentially can be improved with a better attribute predictor which we will consider in our future work.
\begin{figure*}[t]
\centering
%  \subfigure[Market-1501 \label{fig3ta}]{\includegraphics [scale=.531]{ROC_1.eps}}
%  \subfigure[DukeMTMC-ReID\label{fig3tb}]{\includegraphics [scale=.531]{ROC_2.eps}}
%  \subfigure[PETA\label{fig3tc}]{\includegraphics [scale=.531]{ROC_4.eps}}
 \subfigure[Market-1501 \label{fig3ta}]{\includegraphics [scale=.531]{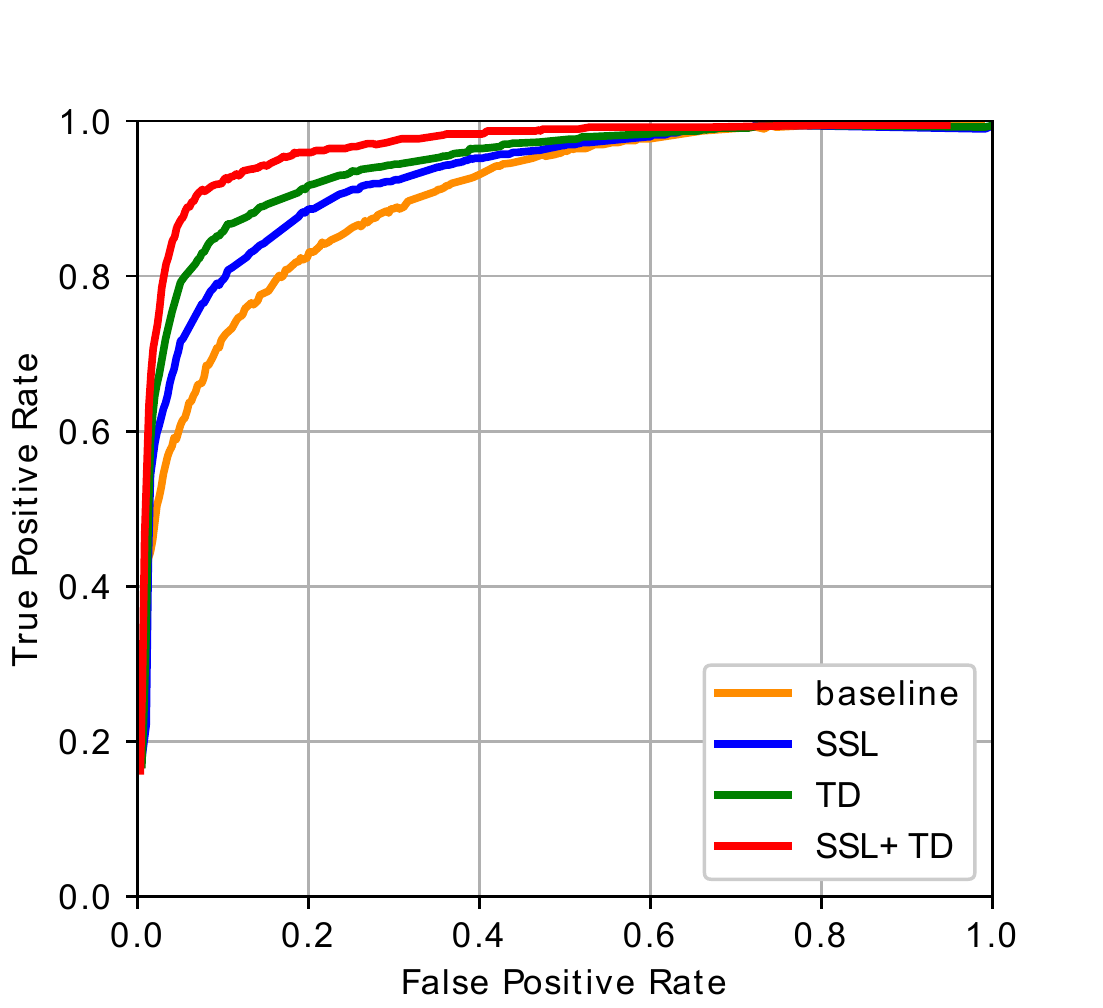}}
 \subfigure[DukeMTMC-ReID\label{fig3tb}]{\includegraphics [scale=.531]{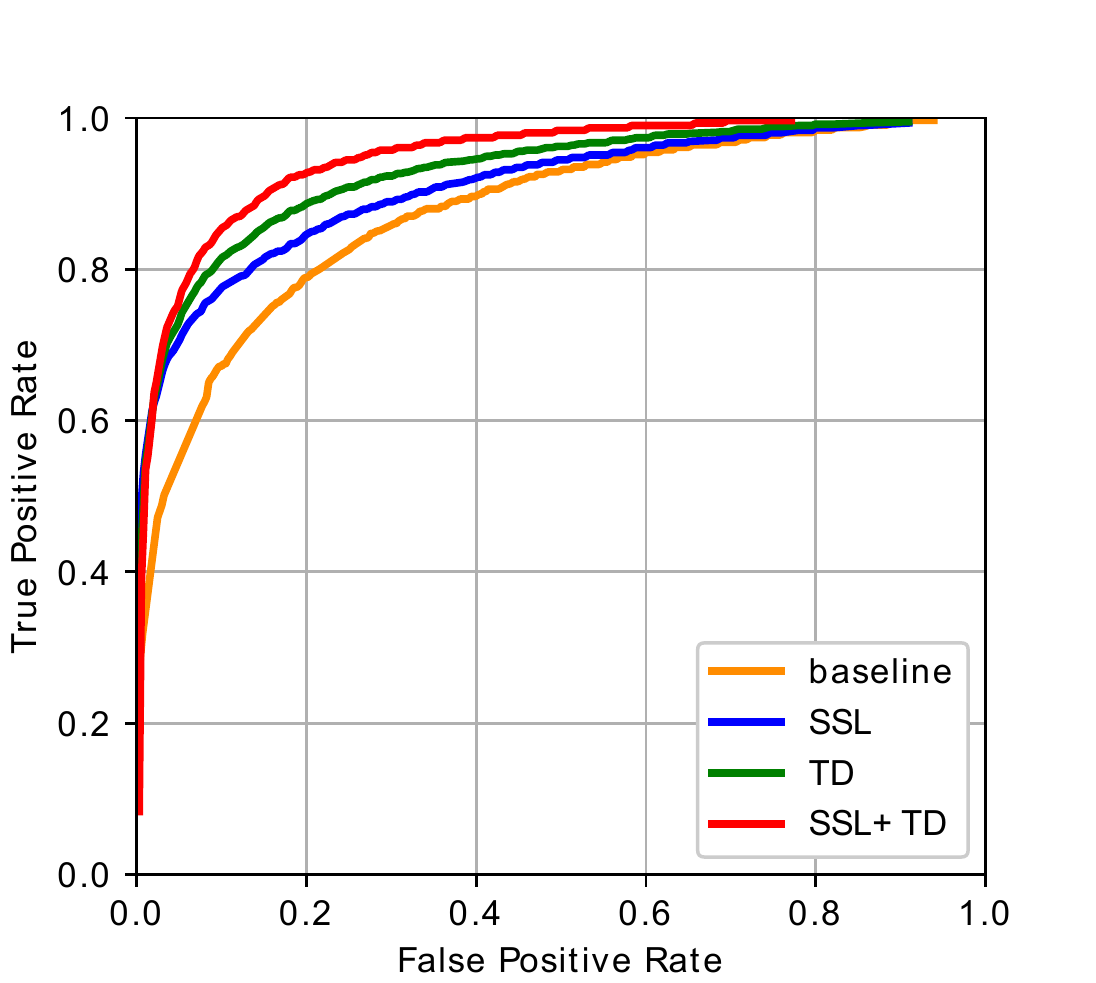}}
 \subfigure[PETA\label{fig3tc}]{\includegraphics [scale=.531]{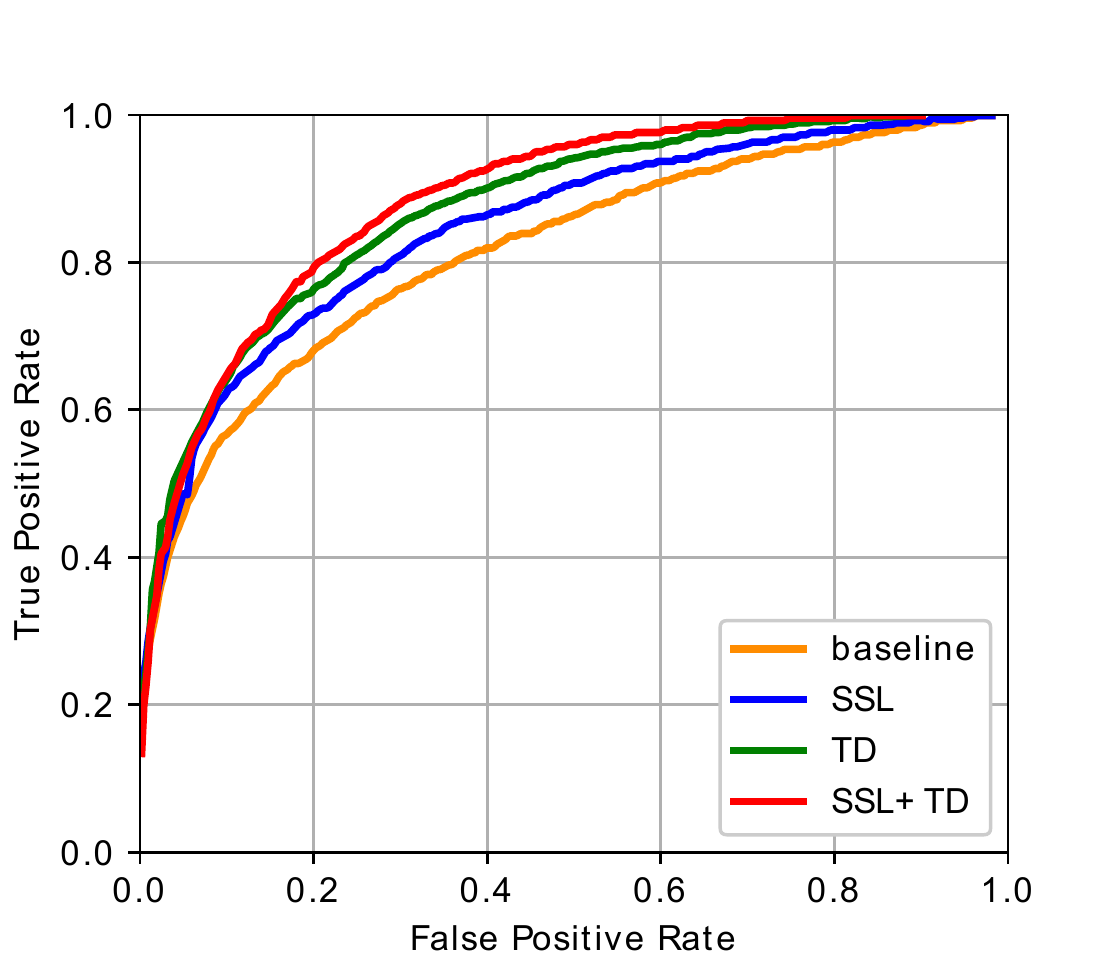}}
\caption{Contributions of the SSL and TD separately and jointly by comparing ROC curves. The baselines are as follows: case (1), the orange curve where the  TD and  SSL  are both ignored  and  the tensor is directly used for ReID, case (2), the blue curve where SSL is applied on the case (1), case (3), the green curve where the TD is applied on the case (1), and case (4), the red curve where  TD and SSL are both applied on the case (1).}\label{fig_4} 
\end{figure*}
\subsection{Baselines and Comparison}
We compare  our proposed model with current state-of-the-art attribute-based person ReID such as \cite{tay2019aanet, lin2019improving} and non-attribute-based person ReID methods in Table   \ref{tab_4} and Table \ref{tab_5}. The methods \cite{tay2019aanet, lin2019improving} are the most recent research work for attribute-based person ReID. APR (w/o attri), and APR (w/o ARM)  are the baselines for APR \cite{lin2019improving} without considering the attributes and  attribute re-weighting module, respectively. Moreover, PCB+RPP \cite{sun2018beyond} is the improved version of the PCB \cite{sun2018beyond} where a refined part pooling is added to the PCB method. Here, we also report the performance of two other baselines in which, for the first baseline (i.e., Baseline (1) in Table. \ref{tab_4} \& Table. \ref{tab_5}), we entirely remove the tensor operator and tackle the problem with our backbone CNN model fine-tuned on the labels of the identities. In the second baseline (i.e., Baseline (2) in Table. \ref{tab_4} \& Table. \ref{tab_5}), we fine-tune our backbone CNN model using labels of the attributes and then we simply use the Euclidean distance between the embedded features  of the query images and the probe images during the retrieval or testing  phase. Comparing the results in Table \ref{tab_4} indicates that our method applied on the Market1501 dataset outperforms the other state-of-the-art method. The rank 1 accuracy and mAP of our method on the Market1501 dataset are $98.38\%$, $90.16 \%$, respectively while the best compared results related to attribute and non-attribute-based methods which are SCSN \cite{chen2020salience}, and AANet \cite{chen2019abd}  are $92.40\%$, $88.30 \%$ and  $93.93\%$, $83.41\%$, respectively.

For the DukeMTMC-ReID dataset, as reported in Table \ref{tab_5}, the rank 1 accuracy and mAP of our method are  $95.42\%$, $84.74\%$ while the best compared results related to attribute and non-attribute-based methods which are Mancs \cite{wang2018mancs}, and AANet \cite{chen2019abd} are $93.10\%$, $82.30\%$, and  $87.65\%$, $74.29\%$, respectively. Furthermore, the results demonstrate that the entire model outperforms the Baseline (1) and Baseline (2) with a  considerable margin.   

In further study, we ablated the SSL term in our loss function (i.e., our method w/o SSL in Table. \ref{tab_4} \& Table. \ref{tab_5}) to see its effectiveness on the performance. In this study, we also train our CNN model in a case where we use the tensor model considering SSL but ignores the TD strategy to reduce the number of the parameters. We call this baseline  w/o TD. Moreover, we study another case where we use the tensor model  but ignoring both the TD and SSL strategies to reduce the number of the parameters. We call this baseline  w/o (TD+SSL).  Here, we used  Algorithm (1) \& (2) presented in Section 5. 2 to estimate the components of the original tensor for ReID using Eq. (8). Comparing the results of these baselines in Table. \ref{tab_4} and Table. \ref{tab_5}  demonstrate that the entire model benefits from reducing parameters based on our SSL and TD methods. 

\begin{figure*}[t]
\centering
\includegraphics[scale=0.47]{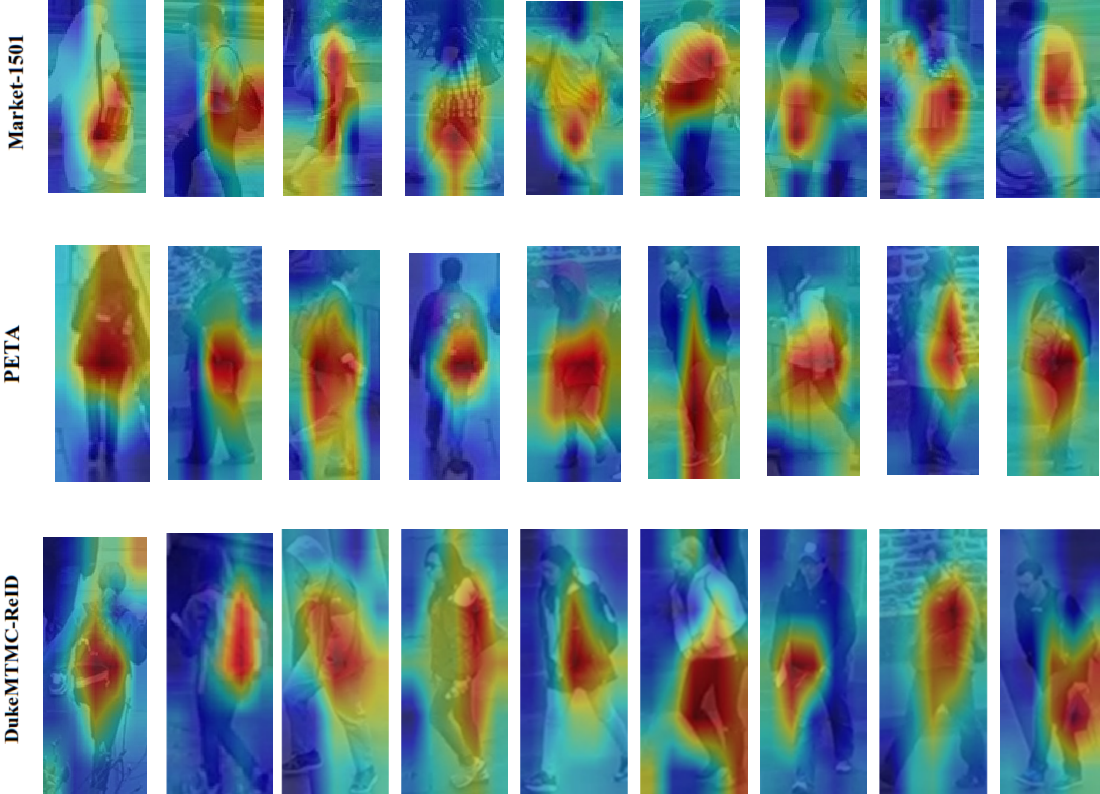}
\caption{Examples of attribute features attention in  our tensor-based ReID model for Market-1501, DukeMTMC-ReID and PETA datasets.}\label{fig_55}
%\vspace{-5.5mm}
\end{figure*}
\begin{figure*}[t]
\centering
 \subfigure[DukeMTMC-ReID\label{fig3ta}]{\includegraphics [scale=.365]{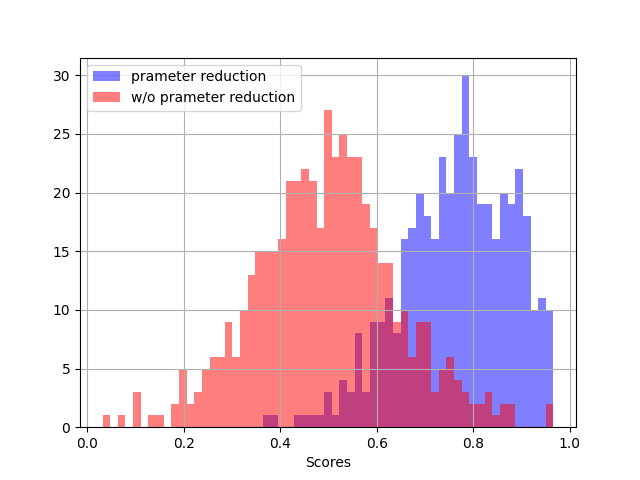}}
 \subfigure[PETA \label{fig3tb}]{\includegraphics [scale=.365]{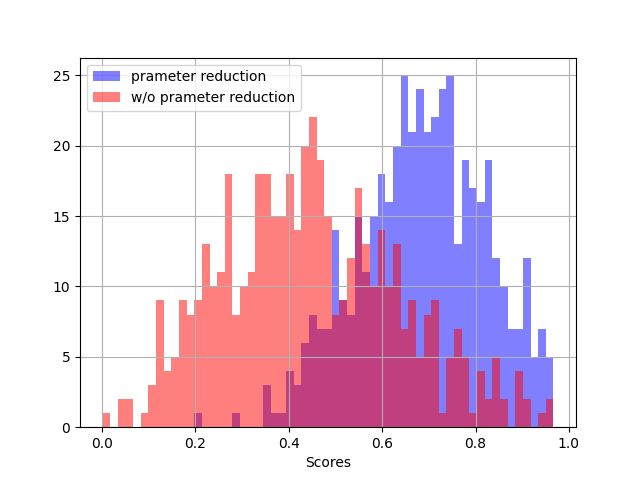}}
 \subfigure[Market1501 \label{fig3tc}]{\includegraphics [scale=.365]{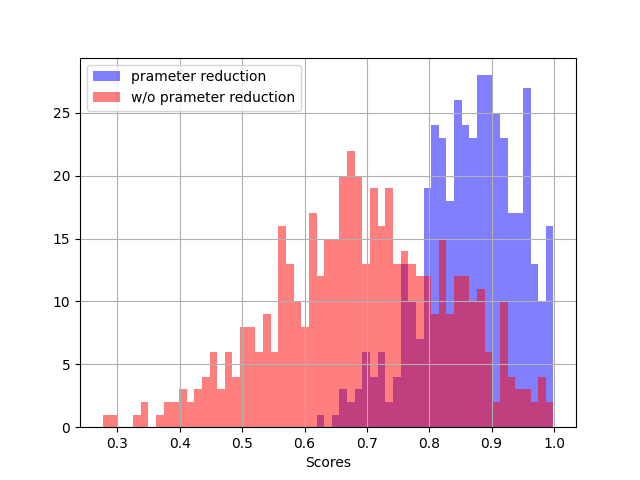}}

\caption{Distribution histogram of the scores for two cases of our tensor-based ReID method: case (1), the blue histogram where the  parameters of the tensor are reduced by TD and SSL, and case (2), the red histogram where the tensor is directly used without reducing the parameters.}\label{fig_7} 
\end{figure*}
\begin{figure*}[t]
\centering
%  \subfigure[DukeMTMC-ReID\label{fig3ta}]{\includegraphics [scale=.365]{score_1.eps}}
%  \subfigure[PETA\label{fig3tb}]{\includegraphics [scale=.365]{score_2.eps}}
%  \subfigure[Market1501 \label{fig3tc}]{\includegraphics [scale=.365]{score_3.eps}}
 \subfigure[DukeMTMC-ReID\label{fig3ta}]{\includegraphics [scale=.365]{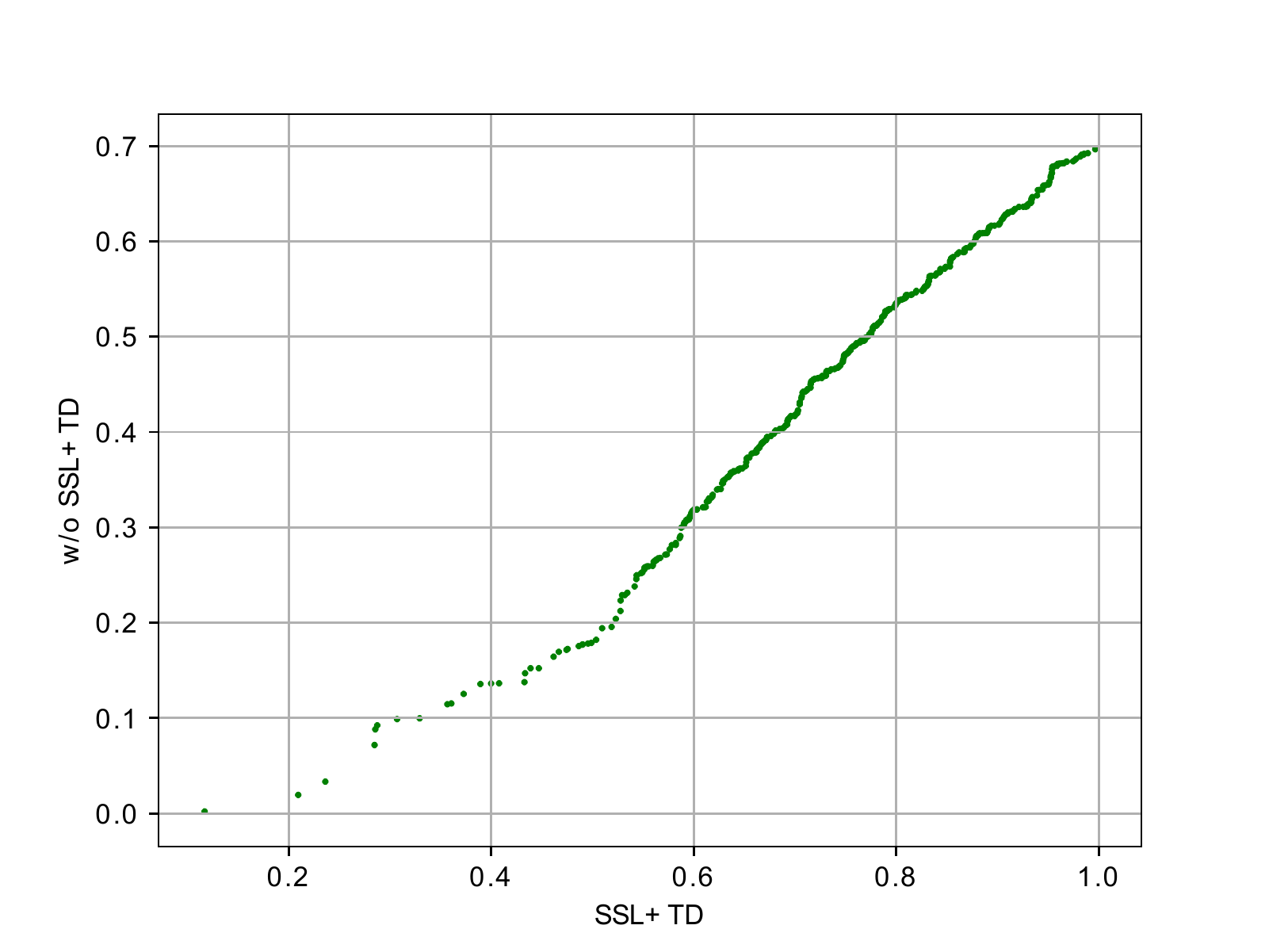}}
 \subfigure[PETA\label{fig3tb}]{\includegraphics [scale=.365]{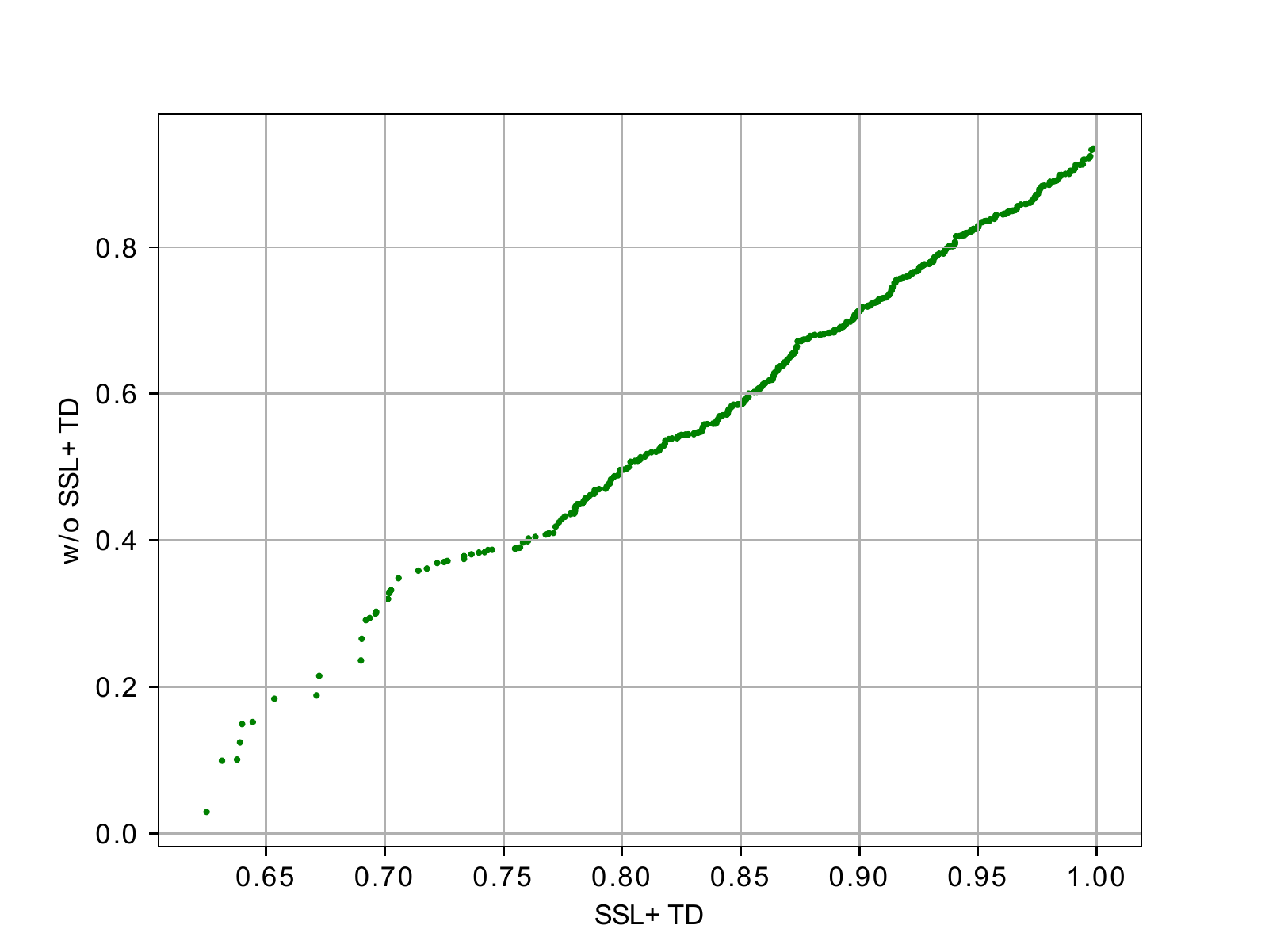}}
 \subfigure[Market1501 \label{fig3tc}]{\includegraphics [scale=.365]{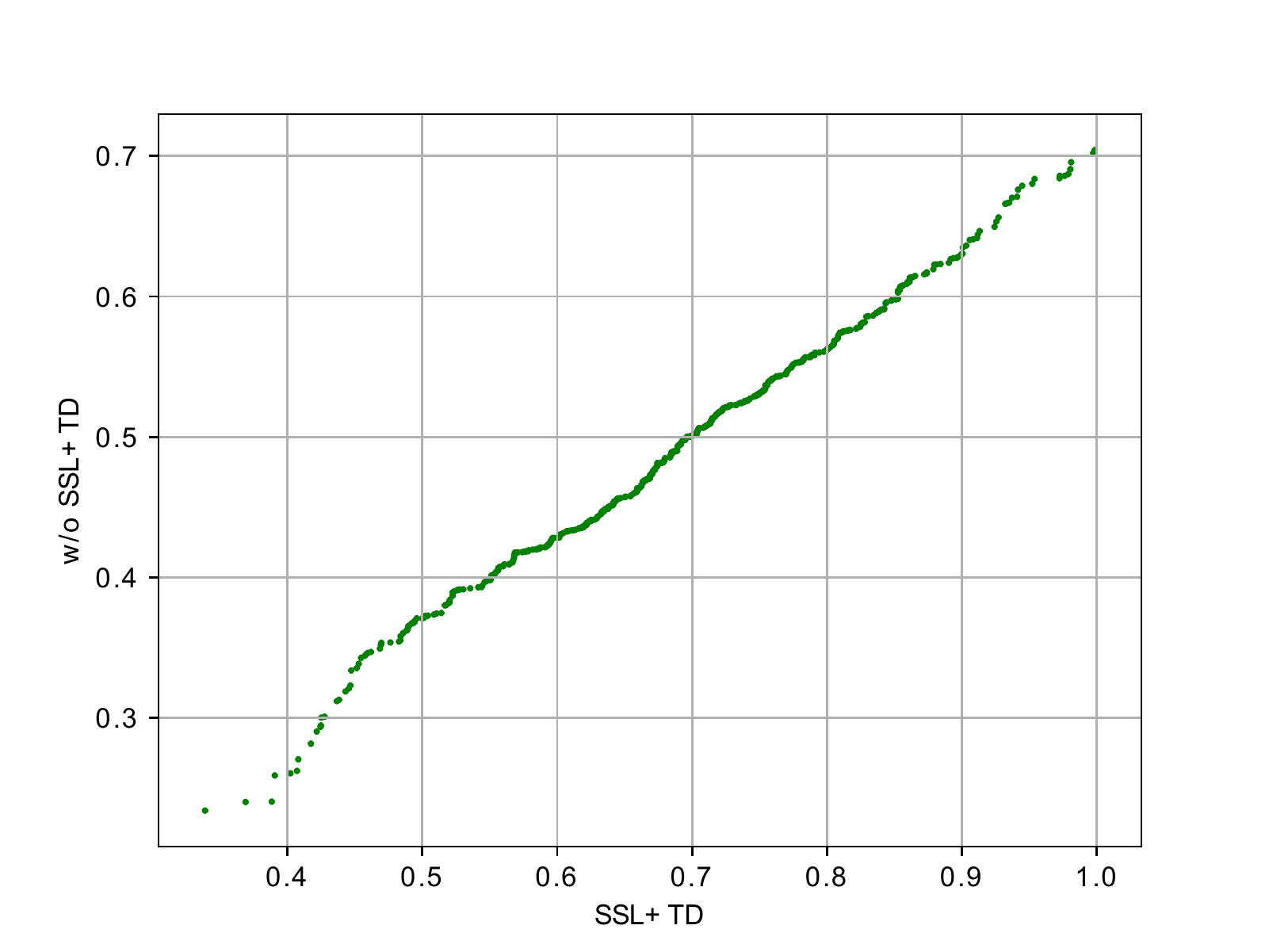}}

\caption{Figures indicate that how the tensor-based identifier which ignores the TD and SSL regularization mimics the scores of the identifier which uses both the TD and SSL regularization techniques. The scores are obtained from 500 positive pairs identified correctly by both identifiers.}\label{fig_8} 
\end{figure*}
\subsection{Further Analysis: Contributions  of the SSL and TD}
In this section, we further study the effect of the SSL and TD on the overall ReID performance in  our tensor-based model. In this study, we consider four different scenarios. In the first  scenario, we ignore both the TD and SSL strategies, and directly use the tensor for ReID; we call this scenario "baseline".  Note that in this case, we used Algorithm (1) \& (2)  to  estimate  the components  of  the tensor for ReID based on Eq. (8). In the second scenario, we only apply the SSL regularization technique on the tensor and then use it for ReID; we call this scenario  "SSL". In the third scenario, we only  apply the TD  technique on the tensor and then use it for ReID; we call this scenario  "TD".  Finally in the fourth scenario, we apply both the SSL regularization and TD techniques on the tensor;  we call this scenario "SSL+TD". In this study, we  train the model on the PETA, Market-1501, and DukeMTMC-ReID datasets for each of four scenarios and report the performance. Here,  we  plotted the receiver operating characteristic (ROC) curve  for the testing set and compared the results for different scenarios. Comparing the ROC curves in Fig. \ref{fig_4} for these scenarios (baseline, SSL, TD, and SSL+TD) demonstrates that the SSL regularization term and TD in our model play an important role on improving the overall ReID performance. Specifically, the ROC curves indicate that adding the SSL constraint on both  the original  and decomposed tensors increases the ReID performance. This observation is evidence for the effectiveness of the SSL regularization term in our model for creating an stable learning problem by reducing the total number of the tensor parameters during the training phase. Moreover, comparing the results obtained from the SSL case with the TD case indicates that the importance of the TD in our model is greater than applying the SSL regularization term in our  model.
\begin{table*}
\centering
\small
\caption{Effect  of  the  SSL  regularization on  the  computation: Level  of  the  speedup  and  percentage  of  the  sparsity  in our model on different datasets. }
\scalebox{0.91}{\begin{tabular}{|c|c|c|c|c|c|c|c|c|c|c|c|c|c|}
 \hline
\multicolumn{1}{|c}{\multirow{1}{*}{}} &\multicolumn{4}{|c}{\multirow{1}{*}{Market-1501}} &\multicolumn{4}{|c|}{DukeMTMC-ReID}&\multicolumn{4}{|c|}{PETA}\\ \cline{2-13}  
\multicolumn{1}{|c}{\multirow{1}{*}{Sparsity level}} &\multicolumn{1}{|c}{\multirow{1}{*}{Top}}&\multicolumn{1}{|c}{\multirow{1}{*}{Front}}&\multicolumn{1}{|c}{\multirow{1}{*}{Side}}&\multicolumn{1}{|c}{\multirow{1}{*}{Speedup}} &\multicolumn{1}{|c}{\multirow{1}{*}{Top}}&\multicolumn{1}{|c}{\multirow{1}{*}{Front}}&\multicolumn{1}{|c|}{\multirow{1}{*}{Side}}&\multicolumn{1}{|c}{\multirow{1}{*}{Speedup}}&\multicolumn{1}{|c}{\multirow{1}{*}{Top}}&\multicolumn{1}{|c}{\multirow{1}{*}{Front}}&\multicolumn{1}{|c|}{\multirow{1}{*}{Side}}&\multicolumn{1}{|c|}{\multirow{1}{*}{Speedup}}\\ \hline \hline
Values & $17.48 \%$ &$20.18  \%$ &$18.23\%$ &$\times 3.4$&$22.61 \%$&$25.15\%$&$19.87 \%$ &$\times 4.1$ &$24.67 \%$& $22.68  \%$ &$20.45 \% $&$\times 3.9$\\ \hline
\end{tabular}}
\label{tab_6}
\end{table*}
\begin{figure*}[t]
\centering
\includegraphics[scale=0.4]{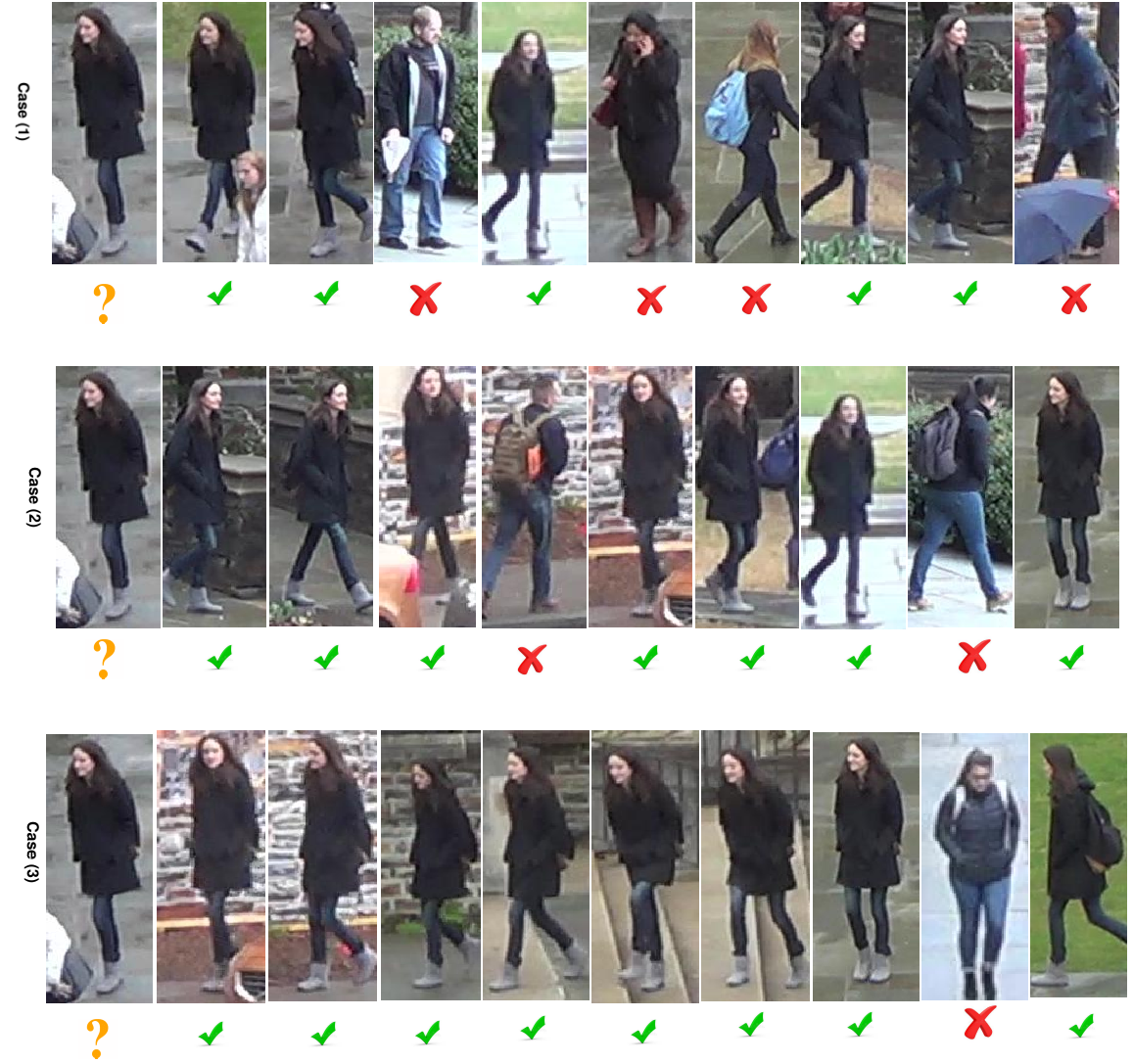}
\caption{Qualitative results: retrieved images from three cases.  case (1) where we entirely remove the tensor operator and simply use the CNN model trained on the attributes labels, case (2) where  we  use our backbone  CNN  trained on  the identities labels, case (3) where we use tensor-based  model to fuse attribute  and  identity  features for  ReID. }\label{fig_6}
%\vspace{-5.5mm}
\end{figure*}

\subsection{Attention of the Attributes Features for ReID}
Inspired by  \cite{zhou2016learning}, we deploy a
Class Activation Map (CAM) to expose the implicit attention of the attribute features on the images  during the person ReID.  We  can learn that information related to the attributes in the images is triggered by semantic regions in the images when we impose it in our tensor model for ReID. From the CAM in Fig. \ref{fig_55}, we can observe that the attribute network provides a "visual explanation" for the region that it concentrates on during the person ReID.
 CAM indicates that the attribute network localizes important regions of the image related to the attributes  to provide complementary information for ReID. In other words, the attribute network determines the regions of the image which are relevant to attributes information.
 
 In our model,  a global average pooling operation is performed on the feature maps obtained from the final convolutional layer, and then a linear  layer for each attribute is conducted on the GAP outputs  to determine if such an attribute is present in the image. For example, in ResNet-50, the last convolutional layer has $2,084$ filters. For a $224\times 224$ input image, the output shape of the last convolutional layer is $2,084\times7\times 7$ due to applying max-pooling operations from previous layers. For each of $2,084$ channels, we have a $7 \times 7$ spatial mapping resolution. The GAP layer  takes each of these $2,084$ channels and returns their spatial average. The parameters associated with a particular attribute predictor assigns a weight to each elements of the GAP layer output. For each of the attributes, these weights represent the significance of each of the channels in a way that the channels with high activation will have larger weights for localizing that attribute in the image. 
% In order to obtain the class activation map, we plug $G_k=\sum_{x,y}g_k(x,y)$ into the class score $S_c$ as defined in Eqs (1) and (2). Thus we have
% \begin{equation}
%   S_c=\sum_kw_k^c\sum_{x,y}g_k(x,y)=\sum_{x,y}\sum_k w_k^c g_k(x,y).  
% \end{equation}
% We declare $A_c$ as the CAM for facial attribute $c$ such that each spatial grid is determined as follows:
% \begin{equation}
%     A_c(x,y)=\sum_k w_k^c g_k(x,y).
% \end{equation}

% Therefore, we can write $S_c=\sum _{x,y}A_c(x,y)$ and in conclusion $A_c(x,y)$ directly represents the significance of the activation at each spatial element $ (x,y)$ which results in  prediction of the attribute $c$ in a given image. The CAM is directly obtained by a linear weighted sum of  visual patterns (i.e., $\{g_1,..,g_K\}$, where, $K$ is the number of feature maps) at different locations. 
% Finally, we simply upsample the CAM to  the input image size by using nearest interpolation method. By performing this, we can recognize the image regions which are most related to a particular facial attribute. The results in Fig. \ref{gradcam} demonstrate the output of the Grad-CAM on the images from DukeMTMC-ReID dataset.
\subsection{SSL and Speedup}
The SSL regularization technique attempts to zero out the groups of  weights related to the redundant slides in the tensor during the training. This strategy reduces the total number of  parameters which ultimately results in speedup in computation as well. We study the effect of the SSL  on the computation and report the level of the speedup and  percentage of  the sparsity when using SSL compared to the case where the SSL regularization term is disregarded.  Table \ref{tab_6}  report the GPU results  (GeForce GTX TITAN X 12GB) on the Market-150, DukeMTMC-ReID, and PETA datasets. Here, we report the percentage of the sparsity on each mode of the tensor (top, side, front), and the total level of speedup. The results indicate that the average percentage of the sparsity  in our sparse tensor-based model trained on the Market-150, DukeMTMC-ReID,  and PETA are $18.63\%$, $22.54 \%$, $22.6 \%$, respectively, and the obtained speedup on these datasets are $\times 3.4$, $\times 4.2$, and $\times 3.9$, respectively.
\subsection{Level of Confidence for ReID using our Model}
In this section, we aim to investigate the level of our model confidence on the true positive pairs in ReID. Specifically, our goal is to study whether or not parameter reduction has any positive impact on the confidence of the ReID.  To conduct this study, we consider two cases where we ignore  the SSL and TD used for parameter reduction as well as  the case where we consider both of them to reduce the number of parameters for ReID. We conducted this study on the Market-1501, DukeMTMC-ReID, and PETA. Specifically,  for each of these datasets, we selected 500 positive pairs identified  correctly by both the aforementioned cases, and we plotted the distribution of their scores during the ReID in Fig. \ref{fig_7}. In this figure, the red and  blue distributions indicate the first and second case, respectively. Moreover, in Fig. \ref{fig_8}, we also plotted the values of the identification scores for these 500 pairs using the two cases to investigate how the scores from the ReID baseline trained based on the first case essentially deviate from that of the ReID baseline trained based on the second case. As it can be observed from  Fig. \ref{fig_7} and Fig. \ref{fig_8}, we can learn that using SSL and TD strategies for reducing the number of the model parameters results in increasing our tensor-based model confidence during the ReID. 
\subsection{Qualitative Results}

Here, we aim to qualitatively investigate the feature representation of our tensor-based model for ReID. In this study, we consider three cases for feature representation. For each of these cases, we simply use the Euclidean distance to rank the similarity between the features  of the query images and  the probe images. In the first case, we  entirely remove the tensor operator and simply use our backbone CNN model fine-tuned on the labels of the attributes for ReID. In the second case,  we perform person ReID based on our backbone CNN model fine-tuned on the labels of the identities. In the third case, we consider our tensor-based model in which both the attribute and identity features are considered for the ReID.  

For qualitative comparison, we randomly select a query image from  the DukeMTMC-reID dataset and rank the testing images based on the similarity using these three cases, and then choose the first nine images shown in Fig.  \ref{fig_6}. In Fig. \ref{fig_6}, we can observe that case 3, case 2, and case 1 retrieve more accurate and relevant images from the probe during testing, respectively. Specifically, this qualitative result indicates that the features obtained from the second case (identity features) can better preserve the semantic similarity between the images in the embedding feature space compared to the first case (attribute feature). Moreover, comparing the first and second cases with the third case, where we integrate both the identities and attributes features using the tensor, demonstrates that  our tensor-based model  benefits from the attribute features as a source of auxiliary information for the ReID. Moreover, from Fig. \ref{fig_6} we can observe that the retrieved images using the third case are more consistent with each other compared to other two cases which potentially indicates the efficacy of our tensor-based model in representing the images in the embedding feature space. 
\subsection{Limitation, discussion and Future Work}
 The major issue in our method  is that it  requires datasets annotated with attributes. Many of the person ReID benchmarks are not annotated with attributes, and annotating the dataset with attributes manually, similar to the method \cite{lin2019improving}, is tedious and labor intensive. However, recent research in semi-supervised learning \cite{sohn2020fixmatch}, and self-supervised learning based on CNN \cite{chen2020simple,chen2020improved} have provided promising results for standard tasks such as image classification. Thus, one of the potential solutions for this issue in our  ReID method is to partially label the datasets with the attributes \cite{lin2019improving} and then use the unlabeled data via advanced semi-supervised or self-supervised learning methods to represent attribute-based features properly in our framework. Here, we represent attribute features using self-supervised learning paradigm. 
 
 In this learning paradigm, we do not need a large amount of images labeled by attribute labels. Instead we pre-train our CNN model using a self-supervised learning paradigm.  Here, we use the MoCo v2 self-supervised learning \cite{chen2020improved}, as it is a strong and efficient self-supervised learning method. Specifically, we use SimCLR \cite {chen2020simple} style data augmentation  for the unlabeled images in the contrastive loss, and follow the implementation details in MoCo v2  where we use a two-layer MLP on the top of the last feature layer to map image features to 128 dimensions, and then use a momentum updated model to calculate the key features in the memory bank. Here, we stored $128$ mini-batches in the memory and each mini-batch contains $128$ samples then the size of our memory bank is $128 \times 128 = 16,384$. After pre-training the attribute network, we select only a small number of images and then label the selected images with attributes to fine-tune the attribute network.  For this study, we conducted experiments on two publicly well-known person ReID benchmarks namely MSMT17, and CUHK03.

 \textbf {MSMT17} is  a public person Re-ID dataset consisting of  126,441 images with 4,101 identities captured by a 15-camera network, including 12 outdoor and 3 indoor. In this dataset, Faster RCNN [28] has been used to annotate the bounding boxes [28]. This dataset is  large-scale, and contains more complicated and dynamic scene characteristic, which makes it challenging for person ReID.

 \textbf{CUHK03} contains  14,097 images  with 1,467 different identities. In this dataset, images are collected  from six cameras  and each identity is captured by two cameras. This dataset includes fewer samples, and the viewpoint variations and occlusion problems are not negligible, which makes it more challenging  for ReID.

 In this study, we used rank  1  accuracy and mAP metrics to evaluate the performance of our  model when there are few attribute labels available during the training.   We compared the results with recent state-of-the-art ReID approaches. While our method cannot outperform the state-of-art method, it is  comparable with it and still preforms better than many other recent work presented in the literature. For example, the results indicate that our method outperforms IANet \cite{hou2019interaction}, GLAD \cite{wei2017glad}, and PDC \cite{su2017pose} on the MSMT17 dataset, and recent methods including MGN \cite{lin2019improving}, Tricks \cite{luo2019bag},  CASN (PCB) \cite{zheng2019re}  on the  CUHK03 dataset.  Moreover, our results still outperform the baseline (1) in Sec. 8. 6 which indicates that our tensor-based model can benefit from the self-supervised  learning paradigm used for learning the attribute features. Here, in this study, we used  the set of 23 attributes used for annotating the DukeMTMC-ReID dataset. Here, we selected 250 images from each of the MSMT17 and CUHK03 datasets and manually annotated their attributes and used the labels for fine-tuning the attribute network. 
 
 \begin{table}[t]
\centering
\caption{Comparison with ReID methods on MSMT17 dataset.}
\scalebox{1}{\begin{tabular}{c| c c c}
 \hline
\multicolumn{1}{c|}{Method} &\multicolumn{1}{c}{Backbone} &\multicolumn{1}{c}{mAP}&\multicolumn{1}{c}{rank 1}\\ [0.5ex] 
 \hline \hline
\textbf{Our method}  & ResNet50&$50.14$&$77.82$\\
Our method w/o SSL  & ResNet50&$48.91$&$75.82$\\
Our method w/o TD & ResNet50&$46.88$&$73.90$\\
Our method w/o (TD+SSL) & ResNet50&$44.45$&$70.95$\\
Baseline (1) & ResNet50&$39.96$&$65.56$\\
\hline
SCSN (4 stages) \cite{chen2020salience} & ResNet50&$58.50$&$83.80$\\
SCSN (3 stages) \cite{chen2020salience} & ResNet50 &$58.00$&$83.0$\\
ABDNet \cite{chen2019abd} & ResNet50 &$\textbf{60.80}$&$\textbf{82.30}$\\
BFE \cite{dai2019batch}& ResNet50 &$51.50$&$78.80$\\
IANet \cite{hou2019interaction} & ResNet50 &$46.80$&$75.50$\\
GLAD \cite{wei2017glad} & ResNet50 &$34.00$&$61.40$\\
PDC \cite{su2017pose} & GoogLeNet &$29.70$&$58.00$\\

%\vspace{-12mm}
\end{tabular}}
\label{table2}
\end{table}

\begin{table}[t]
\centering
\caption{Comparison with ReID methods on CUHK03 dataset.}
\scalebox{1}{\begin{tabular}{c| c c c}
 \hline
\multicolumn{1}{c|}{Method} &\multicolumn{1}{c}{Backbone} &\multicolumn{1}{c}{mAP}&\multicolumn{1}{c}{rank 1}\\ [0.5ex] 
 \hline \hline
\textbf{Our method}  & ResNet50&$73.24$&$78.32$\\
Our method w/o SSL  & ResNet50&$71.94$&$76.55$\\
Our method w/o TD  & ResNet50&$70.53$&$75.09$\\
Our method w/o (TD+TD)  & ResNet50&$68.86$&$72.14$\\
Baseline (1) & ResNet50&$60.28$&$66.85$\\
\hline
SCSN (4 stages) \cite{chen2020salience} & ResNet50&$\textbf{84.00}$&$\textbf{86.80}$\\ 
SCSN (3 stages) \cite{chen2020salience} & ResNet50 &$83.30$&$86.30$\\ 
Pyramid \cite{zheng2019pyramidal} & ResNet101 &$76.90$&$ 78.90$\\
MHN (PCB) \cite{chen2019mixed} & ResNet50 &$72.40$&$77.20$\\ 
BFE \cite{dai2019batch}& ResNet50 &$76.70$&$79.40$\\ 
MGN \cite{lin2019improving}& ResNet50 &$67.40$&$68.00$\\  
CASN (PCB) \cite{zheng2019re} & ResNet50 &$68.00$&$73.70$\\ 
PCB+RPP \cite{sun2018beyond} & ResNet50 &$57.50$&$63.70$\\ 
Tricks \cite{luo2019bag} & SEResNet101 &$70.40$&$72.00$\\ 
Mancs \cite{wang2018mancs} & ResNet50 &$63.90$&$69.00$\\ 
SVDNet \cite{sun2017svdnet} & ResNet50 &$37.80$&$40.90$\\ 
PAN \cite{zheng2018pedestrian}& ResNet50 &$35.00$&$36.90$ 
%\vspace{-12mm}
\end{tabular}}
\label{table2}
\end{table}

\section{Conclusion}
In this study, we proposed a new  method that uses features from human attributes as a source of complementary information for person ReID. Our model uses a tensor to non-linearly  fuse  identity  and  human attributes features and then encourages the parameters of the tensor to represent discriminative fused features using a classification and contrastive learning paradigm. However, our tensor-based method contains a large number of parameters which possible make the training step unstable.  To address this problem, we use Structural Sparsity Learning (SSL) and Tensor Decomposition (TD) methods to reduce the number of  parameters in the tensor in order to create  an  accurate  and  stable  learning  problem. In this work, SSL is applied on the slices of the tensor in each mode where each slice of the tensor in front, top and side modes is considered as a group of weights. This SSL  regularization technique  zeros  out  the  groups  of the weights related to the redundant slides in the tensor during the training, which ultimately  results  in  speedup  in  computation as well. Here, the tensor is an operator that performs two tasks including features fusion and re-identification simultaneously.  Experimental results on the person re-identification benchmarks indicate the effectiveness of our proposed approach. In this work, we have been naive in attribute prediction, which implies that ReID performance potentially can be improved with a better attribute predictor which we will consider in  our future work.  Our experimental results show that the entire model benefits from the SSL regularization and TD to  create a stable learning  problem  by  reducing  the  total  number  of   parameters in the tensor during the training, which results in improving the ReID performance. Other than this advantage, the experimental results also indicate that   using these SSL and TD  strategies  for  reducing  the  number  of  the  model  parameters results in increasing the model confidence during the ReID. Finally, we evaluate our tensor-based model on the benchmarks which are not annotated by attribute labels. Our experimental results indicate that our model potentially still works  well by leveraging a self-supervised learning paradigm to  represent attribute-based features properly for our tensor-based ReID framework.  

% \begin{figure*}[t]
% \centering
% \includegraphics[scale=0.65]{Retr.png}
% \caption{Example of Grad-CAM for visualizing the fused features on  DukeMTMC-ReID dataset.}\label{gradcam}
% %\vspace{-5.5mm}
% \end{figure*}

{\small
\bibliographystyle{IEEEtran}
\bibliography{egbib}
}

\end{document}